\documentclass{article}
\PassOptionsToPackage{numbers, compress}{natbib}

\usepackage[preprint]{main}

\usepackage[utf8]{inputenc} 
\usepackage[T1]{fontenc}    
\usepackage{hyperref}       
\usepackage{url}            
\usepackage{booktabs}       
\usepackage{amsfonts,amssymb,bm,mathtools}    
\usepackage{nicefrac}       
\usepackage{microtype}      
\usepackage{graphicx}
\usepackage{subcaption}
\usepackage{fontawesome}
\usepackage{tikz}
\usepackage{tikz-3dplot}
\usetikzlibrary{arrows.meta}

\usepackage[capitalize,noabbrev]{cleveref}
\hypersetup{
    colorlinks=true,    
    linkcolor=red,      
    citecolor=blue,     
    urlcolor=blue       
}

\usepackage{xcolor}
\usepackage{xspace}

\newcommand{\name}{\textit{Swift}\xspace}

\title{\name: An Autoregressive Consistency\\Model for Efficient Weather Forecasting}

\author{Jason Stock$^1$, Troy Arcomano$^{1,2}$, Rao Kotamarthi$^1$\\
$^1$Argonne National Laboratory $^2$Allen Institute for AI}

\begin{document}

\maketitle

\begin{abstract}
Diffusion models offer a physically grounded framework for probabilistic weather forecasting, but their typical reliance on slow, iterative solvers during inference makes them impractical for subseasonal-to-seasonal (S2S) applications where long lead-times and domain-driven calibration are essential. To address this, we introduce \name, a single-step consistency model that, for the first time, enables autoregressive finetuning of a probability flow model with a continuous ranked probability score (CRPS) objective. This eliminates the need for multi-model ensembling or parameter perturbations. Results show that \name produces skillful 6-hourly forecasts that remain stable for up to 75 days, running $39\times$ faster than state-of-the-art diffusion baselines while achieving forecast skill competitive with the numerical-based, operational IFS ENS. This marks a step toward efficient and reliable ensemble forecasting from medium-range to seasonal-scales. Code and model weights are available at \faGithub\ \url{https://github.com/stockeh/swift}
\end{abstract}

\begin{figure}[!ht]
  \begin{subfigure}[t]{0.34\textwidth}
    \centering
    \begin{tikzpicture}[
  x=10mm,
  y=cos(30)*10mm,
  z={(0,-sin(30)*10mm)},
]
  \def\nspirals{5} 
  \def\cylrad{1.5} 
  \def\cylht{4}    

\draw[thick,-{Circle[length=3pt]}]
  plot[smooth,samples=50,variable=\t,domain=180:360]
    ({cos(\t)*\cylrad},
     {((\nspirals-1)*\cylht/\nspirals)+(\t-180)*\cylht/(360*\nspirals)},
     {-sin(\t)*\cylrad})
    node[above,font=\small] {noise};

  \foreach \i in {1,...,\numexpr\nspirals-2\relax}{%
    \draw[thick]
      plot[smooth,samples=50,variable=\t,domain=180:360]
        ({cos(\t)*\cylrad},
         {(\i*\cylht/\nspirals)+(\t-180)*\cylht/(360*\nspirals)},
         {-sin(\t)*\cylrad})
    \ifnum\i=\numexpr\nspirals-2\relax
      node[pos=1,below,yshift=2mm,font=\small]{$f_\theta([x_T,\hat{x}],T)$}%
    \fi
    ;
  }%

  \foreach \i in {1,...,\numexpr\nspirals-2\relax}{%
    \draw[blue,thick,densely dashed]
      plot[smooth,samples=50,variable=\t,domain=0:180]
        ({cos(\t)*\cylrad},
         {((\i+0.5)*\cylht/\nspirals)+\t*\cylht/(360*\nspirals)},
         {-sin(\t)*\cylrad});
  }%
  \draw[blue,thick,densely dashed,<-]
    plot[smooth,samples=50,variable=\t,domain=0:180]
      ({cos(\t)*\cylrad},
       {(0.5*\cylht/\nspirals)+\t*\cylht/(360*\nspirals)},
       {-sin(\t)*\cylrad})
       node[below,font=\small] at (\cylrad,0.5*\cylht/\nspirals,0) {data};
       
  \draw[blue,densely dashed,<-,thick]
    (-\cylrad-0.5,0.4,0) node[yshift=-2.5mm,font=\small] {$K$} -- 
    node[midway,rotate=90,yshift=3mm,font=\small] {Autoregressive Step} 
    (-\cylrad-0.5,\cylht,0);

  \draw[<-,thick,yshift=4mm,font=\small]
    (-\cylrad-0.15,\cylht,0) node[xshift=-1.3mm,left,font=\small] {$0$} -- 
    node[above] {PF-ODE} 
    node[blue,below,yshift=-1mm] {$x_0 \rightarrow \hat{x}^\prime$} 
    (\cylrad,\cylht,0) node[right,font=\small] {$T$};
\end{tikzpicture}
    \caption{Conceptual diagram}
    \label{fig:cover:concept}
  \end{subfigure}%
  \hfill
  \begin{subfigure}[t]{0.34\textwidth}
    \centering
    \includegraphics[width=\linewidth]{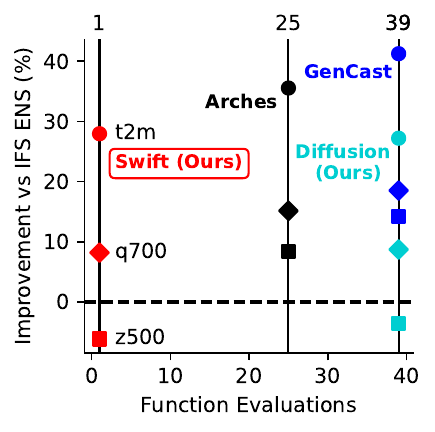}
    \caption{24h forecast error vs compute}
    \label{fig:cover:error}
  \end{subfigure}%
  \hfill
  \begin{subfigure}[t]{0.32\textwidth}
    \centering
    \raisebox{4.5mm}[0pt][0pt]{%
        \resizebox{\linewidth}{!}{%
          \input{figures/cover/illustration.tex}%
        }%
    }
    \caption{Long forecast stability}
    \label{fig:cover:illustration}
  \end{subfigure}
  \caption{Overview of our approach. (a) methodological flow diagram of generating a single-member rollout from noise and an updated conditional state; (b) 39$\times$ faster inference over diffusion baselines by using a single function evaluation with comparable skill to the IFS ENS; and (c) output over a subset of variables shown for a single-member forecast at 24h intervals that is 75 days into the future.}
  \label{fig:cover}
\end{figure}

\section{Introduction}

Recent machine learning models have achieved skill comparable to, and in some cases exceeding, operational weather prediction systems \citep{pathak2022fourcastnet,bi2022pangu,graphcast,nguyen2024scaling}. These deterministic approaches operate at a fraction of the computational cost compared to their numerical counterparts, creating an opportunity to allow for large ensemble systems \citep{lang2024aifsecmwfsdatadriven}. Yet, models such as GraphCast and FourCastNet suffer from a number of problems, including spectral biases \citep{chattopadhyay2023challenges,ebert2025measuring} and insensitivity to initial condition perturbations \citep{butterfly_effect}, limiting their ability to create reliable and calibrated ensembles. Advances in generative modeling, particularly with diffusion models \citep{sohl2015deep,ho2020denoising,song2020score,karras2022elucidating}, offer a grounded formulation to address these problems.

Diffusion-based weather models \citep{price2023gencast,couairon2024archesweather,stock2024diffobs,mardani2025residual,hatanpaa2025aeris} have shown promise in improving small-scale variability, with competitive performance to numerical ensemble systems. However, as they effectively solve differential equations, each forecast step typically makes 20--40 neural function evaluations (NFE), making autoregressive forecasts computationally costly. This compounds with long-lead, seasonal-scale forecasts. To improve efficiency, some works increase their time interval to 12- or 24-hours \citep{price2023gencast,couairon2024archesweather} at the expense of temporal fidelity. Another class of probabilistic weather models combines noise perturbations and multi-model approaches to generate calibrated ensembles \citep{alet2025skillful,lang2024aifs}. For instance, \cite{alet2025skillful} relies on four independent models. While effective, this adds maintenance overhead, limits scalability, and has forecasts that potentially remain prone to instability and artifacts by day 15.

To address these limitations, we introduce \name, a probability flow generative model finetuned autoregressively with a CRPS objective. Our approach builds on a single-step consistency model \citep{lu2024simplifying,song2023consistency}, which preserves many of the favorable properties of diffusion (e.g., learns conditional probabilities, does not require complete state information, and can quantify uncertainty) while being significantly faster. This efficiency not only makes finetuning possible, but consequently yields accurate, well-calibrated ensembles as a result. \name remains effective on medium-range tasks, but extends further into seasonal-scales while preserving temporal fidelity.

In what follows, we summarize the underlying algorithms used to train our models in \cref{sec:background} (including our own baselines). Our method is presented in \cref{sec:method}, where we describe the data and modeling task, our multi-stage training process, network used to achieve stable consistency training, and specific implementation details. This is followed by results in \cref{sec:results}, where we evaluate medium-range forecast skill, assess long-term stability, and present case studies of extreme events and seasonal trends. An overview of our contributions are shown in \cref{fig:cover}.

\section{Background}\label{sec:background}

We first detail the necessary background and diverging modifications of our diffusion (\cref{ssec:diffusion}) and consistency model (\cref{ssec:consistency}), then detail the continuous ranked probability score (\cref{ssec:crps}).

\subsection{Diffusion Models and Flow Matching} \label{ssec:diffusion}

Given clean data $\bm{x}_0\sim p_d$ sampled from our data distribution, diffusion models \citep{sohl2015deep,ho2020denoising,song2020score} diffuse this to noise along $\bm{x}_t = \alpha_t\bm{x}_0 + \sigma_t\bm{z}$ for $t\in[0,T]$ with Gaussian $\bm{z} \sim \mathcal{N}(\bm{0}, \bm{I})$. While there are methods to directly predict the noise, EDM \citep{karras2022elucidating} sets $\alpha_t=1$ and $\sigma_t=t$ to optimize $\mathbb{E}_{\bm{x}_0,\bm{z},t}\left[w(t)\Vert\bm{f}_\theta(\bm{x}_t,t)- \bm{x}_0\Vert^2_2\right]$ with a neural network $\bm{F}_\theta$ preconditioned by $\bm{f}_\theta(\bm{x}_t, t)=c_\mathrm{skip}(t)\bm{x}_t + c_\mathrm{out}(t)\bm{F}_\theta\left(c_\mathrm{in}(t)\bm{x}_t, c_\mathrm{noise}(t)\right)$ and weighting function $w(t)$. The coefficients ensure the objective has unit variance across noise levels. Generating clean data involves iteratively solving the probability flow ODE (PF-ODE) \citep{song2020score} $\tfrac{\mathrm{d}\bm{x}_t}{\mathrm{d}t} = \left[\bm{x}_t - \bm{f}_\theta(\bm{x}_t,t)\right]/t$ from pure noise $\bm{x}_T \sim \mathcal{N}(\bm{0}, T^2\bm{I})$.

Flow matching \citep{lipman2022flow,liu2022flow,albergo2023stochastic,tong2023improving} (also known as rectified flows or stochastic interpolants) considers a linear interpolant with $\alpha_t=1-t$ and $\sigma_t=t$, where the objective is to learn the velocity $\bm{v}_\theta$ using $\mathbb{E}_{\bm{x}_0,\bm{z},t}\left[\Vert\bm{v}_\theta(\bm{x}_t,t)- (\bm{z} - \bm{x}_0)\Vert^2_2\right]$. Sampling involves solving the PF-ODE $\tfrac{\mathrm{d}\bm{x}_t}{\mathrm{d}t} = \bm{v}_\theta(\bm{x}_t,t)$ from $\bm{x}_1 \sim \mathcal{N}(\bm{0}, \bm{I})$ with a classical numerical solver (e.g., Runge-Kutta methods) from $t=1$ to $t=0$.

TrigFlow \citep{lu2024simplifying} unifies EDM and flow matching under a simpler, v-prediction parameterization that is theoretically supported for both diffusion and consistency models (\cref{ssec:consistency}). It satisfies the above unit variance principle and maps arbitrary noise schedules onto a common trigonometric trajectory. Concretely, we considers a spherical interpolant with $\alpha_t=\cos(t)$ and $\sigma_t=\sin(t)$ and noisy sample $\bm{x}_t = \cos(t)\bm{x}_0 + \sin(t)\bm{z}$, where $\bm{z} \sim \mathcal{N}(\bm{0}, \sigma_d^2 \bm{I})$ with the data standard deviation $\sigma_d$, and $t = \arctan(e^\tau/\sigma_d) \in [0, \pi/2]$ with $\tau$ drawn from a prior distribution, such as log-normal or log-uniform. We use the latter in training our TrigFlow models, given our data distribution.

We define our diffusion model $\bm{f}_\theta(\bm{x}_t, t)=\bm{F}_\theta(\bm{x}_t / \sigma_d, t)$ with no further preconditioning, and have the corresponding PF-ODE $\tfrac{\mathrm{d}\bm{x}_t}{\mathrm{d}t} = \sigma_d \bm{F}_\theta({\bm{x}_t}/ \sigma_d, t)$. The training objective under TrigFlow for diffusion estimates the velocity with
\begin{equation}\label{eq:loss.diff}
\ell_{\mathrm{Diff}}\left(\theta\right) = \mathbb{E}_{\bm{x}_0,\bm{z},t} \left[\left\Vert \sigma_d \bm{F}_\theta \left(\tfrac{\bm{x}_t}{\sigma_d}, t\right) - \left(\cos(t)\bm{z} - \sin(t)\bm{x}_0\right) \right\Vert^2_2 \right].
\end{equation}

These formulations, at a high level, are all continuous-time generative models that aim to achieve the same result; while they differ in their parameterizations, they are theoretically equivalent under correct assumptions. However, these methods all share a significant expense in their sampling techniques, whether that be a simpler first-order Euler solver or higher-order method as used in related approaches \citep{karras2022elucidating,price2023gencast}. We consider the latter, though without stochastic Langevin-like churn due to the trigonometric schedule complexity.

\begin{figure}[!t]
    \centering
    \begin{minipage}[b]{0.54\linewidth}
        \centering
        \includegraphics[width=\linewidth]{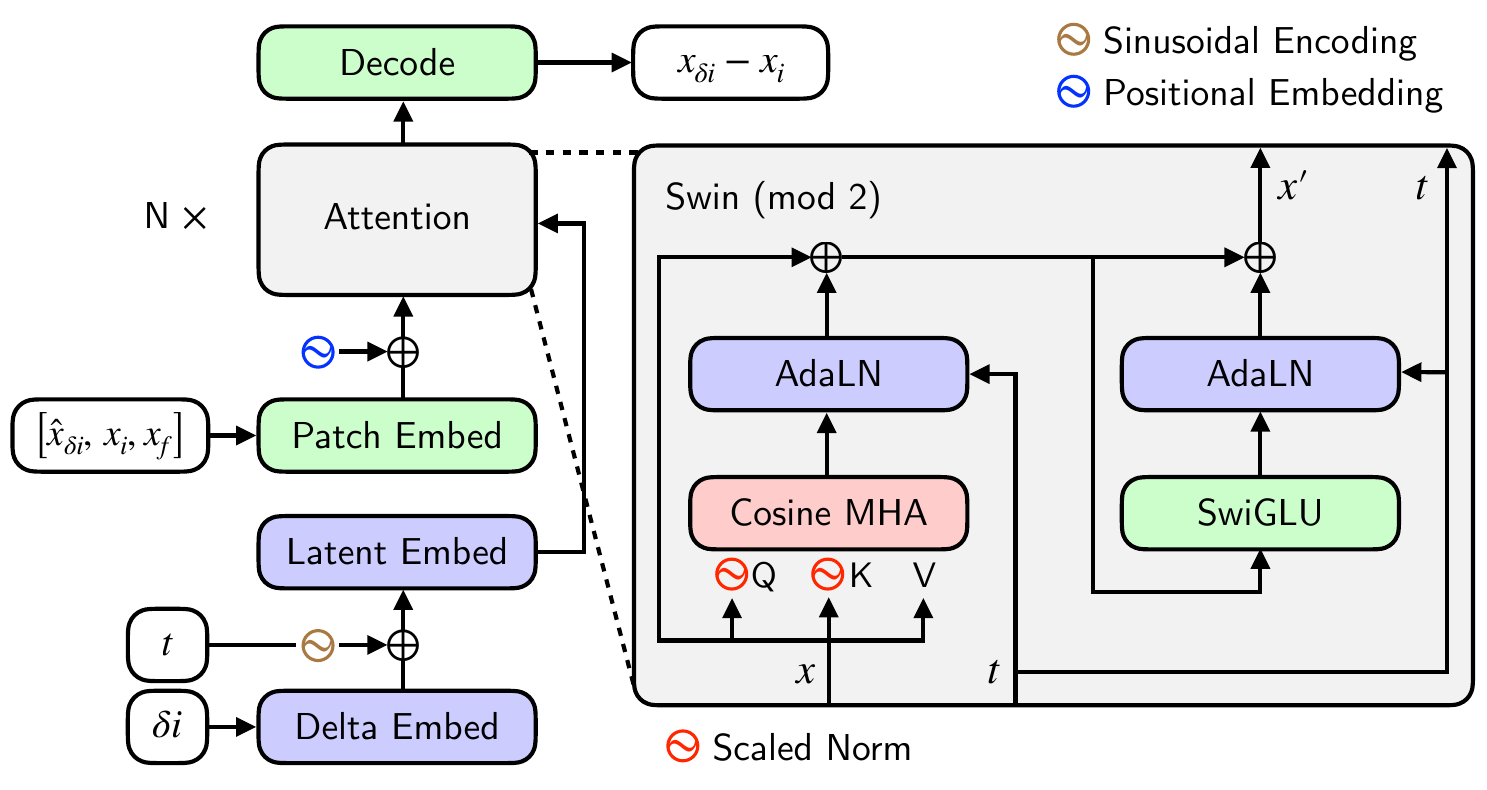}
        \caption{Our proposed network architecture.}
        \label{app:fig:architecture}
    \end{minipage}%
    \hfill
    \begin{minipage}[b]{0.45\linewidth}
        \centering
        \scriptsize
        \captionof{table}{ERA5 surface-level and atmospheric variables (50--1000 hPa) with input forcings.}
        \begin{tabular}{lll}
            \toprule
            Type & Variable & Description\\
            \midrule
             surface & \texttt{t2m} & 2 meter temperature\\
             surface & \texttt{10u} & 10 meter $u$-wind component\\
             surface & \texttt{10v} & 10 meter $v$-wind component\\
             surface & \texttt{mslp} & Mean sea level pressure\\
             atmos. & \texttt{z------} & Geopotential\\
             atmos. & \texttt{t------} & Temperature\\
             atmos. & \texttt{u------} & $u$-wind component\\
             atmos. & \texttt{v------} & $v$-wind component\\
             atmos. & \texttt{q------} & Specific humidity\\
            \midrule
             static & \texttt{z} & Geopotential at surface\\
             static & \texttt{lsm} & Land-sea mask\\
             clock & \texttt{n/a} & TOA incident solar radiation\\
            \bottomrule
        \end{tabular}
        \label{tab:data}
        \vspace{0pt plus 1filll}
    \end{minipage}
\end{figure}

\subsection{Continuous-time Consistency Models} \label{ssec:consistency}

Unlike diffusion, consistency models \citep{song2023consistency,song2023improved,lu2024simplifying} learn to predict clean data $\bm{x}_0$ at the origin of the PF-ODE for any noisy sample $\bm{x}_t$ along the trajectory. Intuitively, this means we need not rely on iterative ODE solvers, and instead can directly estimate data in one (or few) steps. Importantly, this process satisfies the boundary condition $\bm{f}_\theta(\bm{x}, t)\equiv \bm{x}$ when parameterized by $\bm{f}_\theta(\bm{x}_t, t)=c_\mathrm{skip}(t)\bm{x}_t + c_\mathrm{out}(t)\bm{F}_\theta\left(c_\mathrm{in}(t)\bm{x}_t, c_\mathrm{noise}(t)\right)$ with $c_\mathrm{skip}(0) = 1$ and $c_\mathrm{out}(0) = 0$. Under TrigFlow we parameterize the consistency model as a single step solution to our PF-ODE with simplified arithmetic coefficients as
\begin{equation}\label{eq:scm.parameterization}
\bm{f}_\theta(\bm{x}_t, t)=\cos(t)\bm{x}_t - \sin(t)\sigma_d \bm{F}_\theta\left(\frac{\bm{x}_t}{\sigma_d},t\right).
\end{equation}

To learn in the discrete case, the training objective is defined at two adjacent steps with finite distance as $\mathbb{E}_{\bm{x}_t,t}[w(t)d(\bm{f}_\theta(\bm{x}_t, t),\bm{f}_{\theta^-}(\bm{x}_{t-\Delta t}, t-\Delta t)]$, with a weighting function $w(t)$, stop gradient $\theta^-$, finite distance $\Delta t > 0$, and loss measure $d(\cdot,\cdot)$, e.g., $\ell_2$ or Pseudo-Huber \citep{song2023improved}. In the continuous case, \cite{song2023consistency} show when $d(\bm x, \bm y) = \Vert \bm x - \bm y \Vert^2_2$ and we take the limit as $\Delta t \rightarrow 0$, the gradient of the discrete objective is $\nabla_\theta\mathbb{E}_{\bm{x}_t, t}[w(t)\bm{f}_\theta^\top\tfrac{\mathrm{d} \bm{f}_{\theta^-}}{\mathrm{d}t}]$. Observing that $\nabla_\theta \mathbb{E}[\bm{F}_\theta^\top \bm{y}]=\tfrac{1}{2}[\Vert \bm{F}_\theta - \bm{F}_{\theta^-} + \bm{y}\Vert]_2^2$, using an arbitrary vector $\bm y$ independent of $\theta$, we arrive at the following (prior weighting removed) objective
\begin{equation}\label{eq:loss.scm}
\ell_{\mathrm{sCM}}\left(\theta\right) = \mathbb{E}_{\bm{x}_t,t} \left[ \left\Vert \bm{F}_\theta \left(\frac{\bm{x}_t}{\sigma_d},t\right) - \bm{F}_{\theta^-} \left(\frac{\bm{x}_t}{\sigma_d},t\right) - \cos(t) \frac{\mathrm{d}\bm{f}_{\theta^-} \left(\bm{x}_t,t\right)}{\mathrm{d}t} \right\Vert^2_2 \right].
\end{equation}

This function depends on the time derivative, or tangent along our trajectory, of \cref{eq:scm.parameterization} as
\begin{equation}\label{eq:loss.scm.tangent}
\frac{\mathrm{d}\bm{f}_{\theta^-}\left(\bm{x}_t,t\right)}{\mathrm{d}t} = -\cos(t) \left( \sigma_d \bm{F}_{\theta^-}  - \frac{\mathrm{d}\bm{x}_t}{\mathrm{d}t}\right) - r \cdot \sin(t) \left( \bm{x}_t + \sigma_d \frac{\mathrm{d}\bm{F}_{\theta^-}}{\mathrm{d}t} \right),
\end{equation}
where $r$ is a tangent warmup that linearly increases (over the first 3 million training images) to stabilize training and the PF-ODE is approximated by an unbiased estimate $\tfrac{\mathrm{d}\bm{x}_t}{\mathrm{d}t} = \cos(t)z - \sin(t)\bm{x}_0$. We compute $\tfrac{\mathrm{d}\bm{F}_{\theta^-}}{\mathrm{d}t}$ by means of forward-mode automatic differentiation via the Jacobian-vector product in PyTorch and the stopgrad $\theta^-$ by simply detaching the gradient flow. Additionally, following \cite{lu2024simplifying}, we normalize the tangent by the factor $(\Vert \bm y \Vert+c)^{-1}$ with $c=0.1$, but scale $\Vert \bm y \Vert$ to remain invariant to the spatial dimensions.

Note that a pretrained diffusion model defined by \cref{eq:loss.diff} could be distilled into the same or smaller sized model by replacing the PF-ODE with $\tfrac{\mathrm{d}\bm{x}_t}{\mathrm{d}t} = \sigma_d \bm{F}_\mathrm{pretrain}(\tfrac{\bm{x}_t}{\sigma_d},t)$. However, this is out of scope of the current work and we instead consolidate end-to-end training with the consistency model.

\subsection{Continuous Ranked Probability Score (CRPS)} \label{ssec:crps}

Our consistency model is parameterized such that single-step predictions simplify \cref{eq:scm.parameterization} and are given by $\bm{f}_\theta(\bm{x}_t, t)= - \sigma_d \bm{F}_\theta\big(\tfrac{\bm{x}_t}{\sigma_d},t\big)$, assuming we set $t=\tfrac{\pi}{2}$ and sample $\bm{z}_\frac{\pi}{2}\sim \mathcal{N}(\bm{0}, \sigma_d^2 \bm{I})$. The objective in \cref{eq:loss.scm} has no notion of forecast uncertainty and therefore cannot, on its own, ensure ensembles are physically calibrated. However, this simplified parameterization allows us to resample noise to produce diverse ensemble members, which we then can calibrate with the right, domain-driven metric.

CRPS is a strictly proper scoring rule for univariate distributions that quantifies the discrepancy between a forecast cumulative distribution function (CDF) $F$ and an observed value $y \in \mathbb{R}$. This is achieved by integrating $\int_{-\infty}^{\infty} \left(F(z) - \mathbf{1}[z \geq y]\right)^2 \, \mathrm{d}z$, which measures the squared difference between the forecast CDF and a step function at the observation. When the forecast distribution is represented with finite ensemble members, we can approximate $F$ with an empirical CDF. A fair estimate (with the self-comparison removed) \citep{ferro2008effect,ferro2014fair,leutbecher2019ensemble} provides the following unbiased estimate given predictions $\hat{y}$
\begin{equation}\label{eq:crps}
\mathrm{CRPS}\left(\hat{y}^{1:N}, y\right) = \frac{1}{N}\sum_n \lvert \hat{y}^n - y\rvert - \frac{1}{2N(N-1)}\sum_{n\neq n^\prime} \lvert \hat{y}^n - y^{n^\prime}\rvert.
\end{equation}
In this formulation, the first term encourages ensemble members to remain close to the observation, while the second accounts for ensemble dispersion by correcting for self-similarity. This ultimately balances both forecast accuracy and sharpness with ensembles that are well calibrated.

\begin{figure}[!t]
    \centering
    \includegraphics[width=\linewidth]{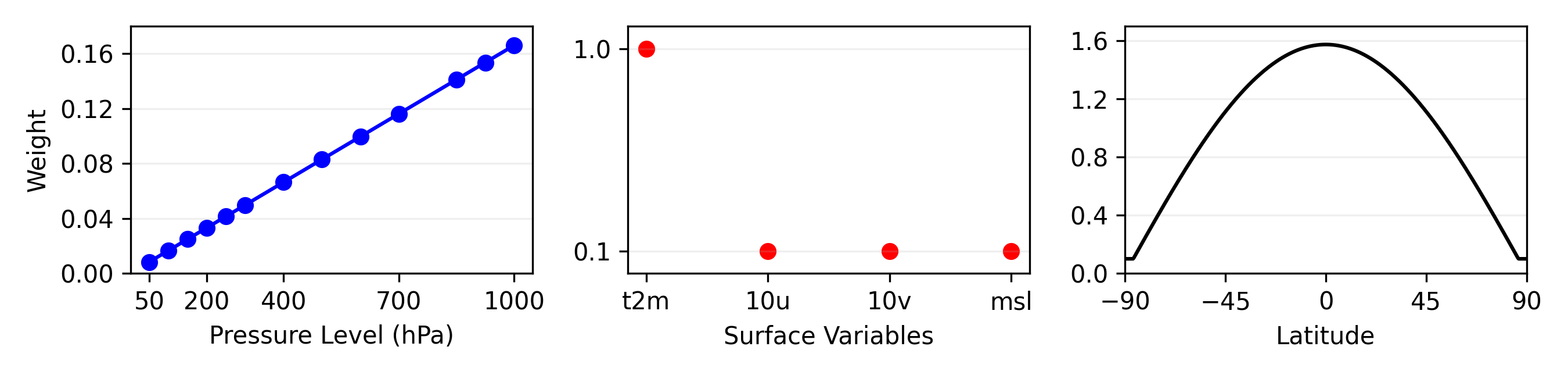}
    \caption{Loss weights used during training. (left) pressure weights applied to atmospheric variables; (middle) surface level weighting to specific variables; and (right) clipped latitude weighting.}
    \label{fig:loss.weights}
\end{figure}

\section{Methodology}\label{sec:method}

Our method connects the concepts in \cref{sec:background} for the task described in \cref{ssec:dataset}. We then introduce our physically motivated training objectives in \cref{ssec:objectives}, followed by our network architecture and modeling specifics in \cref{ssec:architecture} before summarizing implementation details in \cref{ssec:implementation}.

\subsection{Task and Dataset}\label{ssec:dataset}

We model the global evolution of the atmosphere by learning $p\left(x_{i+1} \mid x_{i}\right)$ with the temporal atmospheric states indexed by $i$. These states come from four decades of ERA5 reanalysis data \citep{hersbach2020era5}, as provided by WeatherBench 2 (WB2) \citep{rasp2024weatherbench}. Data are downsampled to $1.40625^\circ$ resolution ($128\times 256$ pixels) at 6-hour intervals, which differs from the native $0.25^\circ$ resolution of our baseline, GenCast \citep{price2023gencast}. Our models use data from 1979--2018 for training, 2019 for validation, and 2020 for testing. 

We predict four surface-level variables (t2m, u10, v10, mslp) and five atmospheric variables (z, t, u, v, q), each at 13 pressure levels ($\{50, 100, 150, 200, 250, 300, 400, 500, 600, 700, 850, 925, 1000\}$ hPa), with additional forcings of top-of-atmosphere solar radiation, surface geopotential, and land-sea mask to stabilize phase shifts during long horizons and to simplify orographic representations (see \cref{tab:data} for more details). To reduce error accumulation, we follow \citep{nguyen2024scaling} and incorporate random dynamic intervals $\delta i \sim \mathcal{U}\{6,12,24\}$ to capture diurnal and synoptic-scale dynamics while preserving temporal fidelity. Each autoregressive step then estimates the residual $x_{\delta i} - x_i$ from the initial state $x_i$ and input forcings $x_f$. A forecast rollout is made for $K$ autoregressive steps at any time delta.

During training and inference, data are z-score standardized using per-variable and per-level statistics from the training set. Residual targets rely on time delta statistics for each interval (approximately Gaussian), while conditional states use the full-field statistics. At each autoregressive step in finetuning and inference, predictions are unstandardized with the correct delta statistics and added back to the initial state $x_i$ to reconstruct the full field $x_{\delta i}$. This becomes the subsequent initial condition. Additional details on the incorporation of conditional states, including the noise level (diffusion time) and data delta time, are provided in \cref{ssec:architecture}.

\subsection{Training Objectives}\label{ssec:objectives}

Our training proceeds in two stages, yielding our final model and two baseline models for comparison. The first stage consists of pretraining a baseline \textit{diffusion} model alongside a base consistency model, \name-\textit{B}. The second stage applies multi-step finetuning to the consistency model, producing  \name, our final calibrated model.

\begin{figure}[!t]
  \centering
  \includegraphics[width=\linewidth]{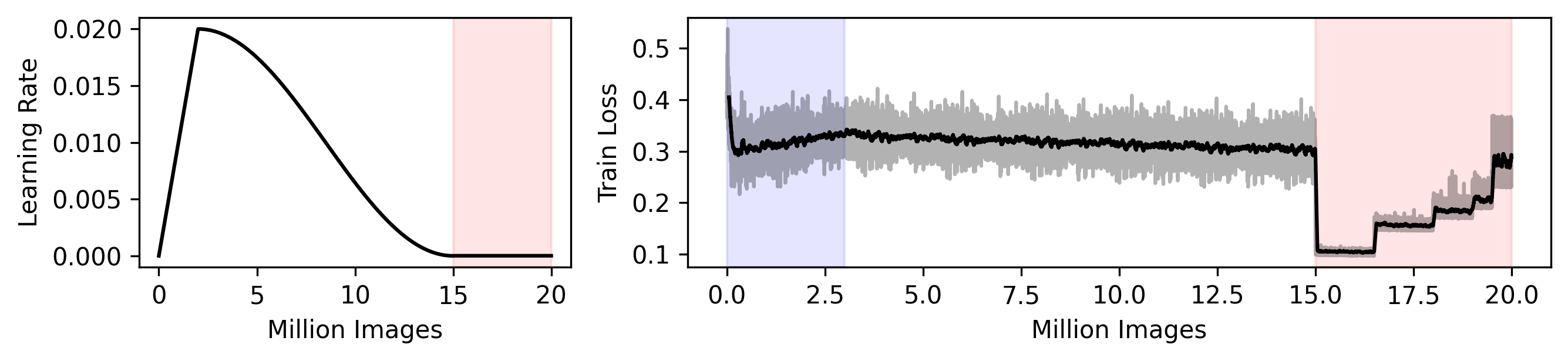}
  \caption{Learning curves for \name. (left) learning rate schedule with Muon $\eta$ during pretraining and AdamW during finetuning (in \textcolor{red!50!white}{red}); and (right) training loss with a 3M tangent warmup (\cref{eq:loss.scm.tangent}, in \textcolor{blue!50!white}{blue}) before multi-step finetuning with $K=1$--$8$ autoregressive steps from 15--20M images.}
  \label{fig:loss.curve}
\end{figure}

\paragraph{Model Pretraining.} We train and formulate both our diffusion and consistency models under TrigFlow (\cref{sec:background}). This v-prediction parameterization allows for simpler, few step samplers and opens the possibility of distilling large-scale diffusion models (e.g. \cite{hatanpaa2025aeris}) into smaller, more efficient models (not shown herein). Such capabilities are not possible with GenCast \cite{price2023gencast} and its EDM formulation. Our pretraining loss includes latitude- and pressure-dependent functions to account for the non-uniform spherical grid and to emphasize near-surface variables \citep{price2023gencast,tong2023improving}. Specifically, the latitude and variable weights, denoted $\alpha(s)$ and $\kappa(v)$ for each variable $v\in\mathcal{V}$ (\cref{fig:loss.weights}), yield
\begin{equation}
\mathcal{L}(\theta) = \frac{1}{|\mathcal{S}|} \sum_{s \in \mathcal{S}} \sum_{v \in \mathcal{V}} \kappa(v) \alpha(s) \ell_{v,s}(\theta),
\end{equation}
where $\mathcal{S}$ is the set of spatial indices over all batches and $\ell$ is either the diffusion or consistency loss in \cref{eq:loss.diff,eq:loss.scm}. Extensive hyperparameter searches led us to adopt Muon \citep{jordan2024muon} for consistency pretraining and AdamW \citep{loshchilov2017decoupled} for diffusion; see \cref{ssec:implementation} for additional details.

\paragraph{Multi-step Finetuning.} Our base, pretrained consistency model (\name-\textit{B}) is not explicitly optimized for long-term atmospheric dynamics, leading to greater error accumulation when applied autoregressively. Considering our v-prediction loss regresses a trajectory-based target at short lead-times, we posit it is misaligned with producing faithful and physically calibrated forecast ensembles. We therefore exploit \name's probabilistic nature and efficient sampling by finetuning on a proper scoring rule, specifically optimizing the unbiased and fair CRPS (\cref{eq:crps}) with weights as
\begin{equation}\label{eq:loss:crps}
\mathcal{L}_\mathrm{CRPS} = \frac{1}{|\mathcal{S}|} \sum_{s \in \mathcal{S}}\sum_{v \in \mathcal{V}} \kappa(v) \alpha(s) \mathrm{CRPS}\left(\hat{\bm{y}}_{v,s}^{1:2}, \bm{y}_{v,s}\right),
\end{equation}
where $\hat{\bm{y}}$ is computed in a single-step from the initial condition and varying Gaussian noise for $N=2$ ensembles and per-batch $\delta i$. We parallelize CRPS with a vector map over batch elements and use sequential gradient checkpointing \citep{chen2016training} over $K=2$ autoregressive steps to allow for higher step counts when needed. Following a curriculum schedule, we train with $K=\{1,2\}$ for 1.5 million images (mimg), $K=3$ for 1 mimg, and $K=\{4,8\}$ for 0.5 mimg, backpropogating in time through $K$ with the loss computed on the final step. This would otherwise be prohibitively costly in time and space for diffusion models, thereby positioning this work among the first to make multi-step finetuning tractable for generative forecasting.

\begin{figure}[!t]
  \centering
  \includegraphics[width=\linewidth]{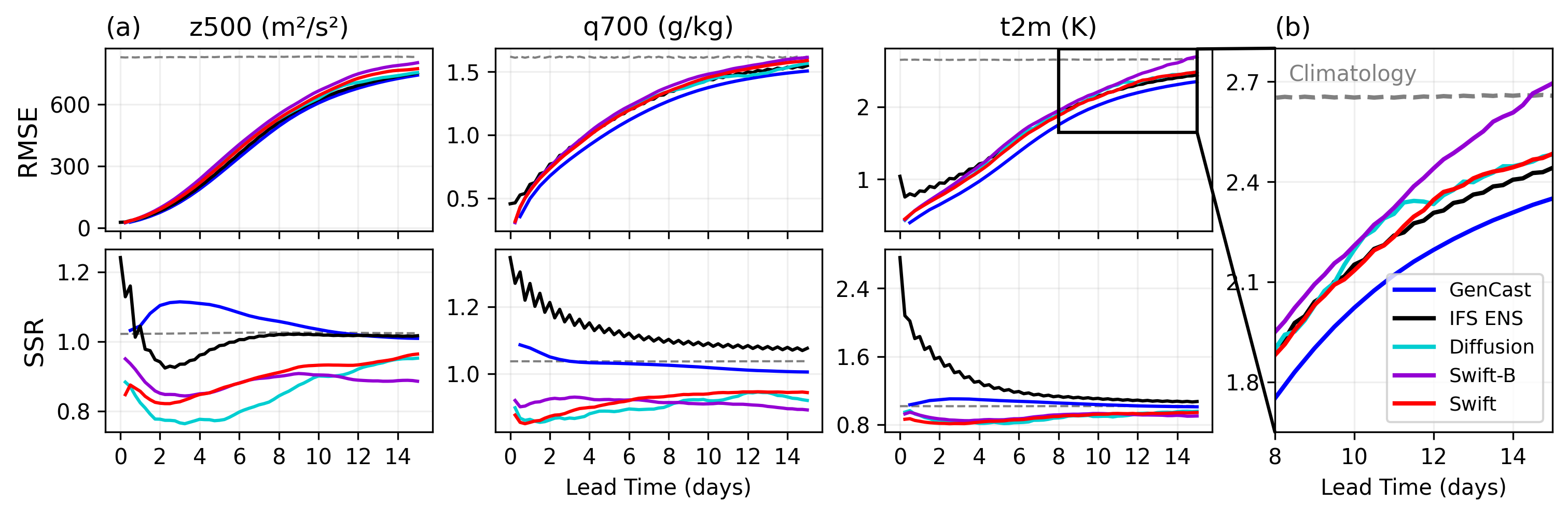}
  \caption{Global forecast skill (on a subset of initials with $\delta i=6$). (a) latitude-weighted ensemble RMSE and spread/skill compared to baselines; and (b) close view benefit of multi-step finetuning.}
  \label{fig:skill}
\end{figure}

\subsection{Model Architecture} \label{ssec:architecture}

We introduce a conditional non-hierarchical Swin transformer (\cref{app:fig:architecture}), extending \citep{liu2022swin,willard2025analyzing} that empirically yields stable training for both diffusion and consistency models (\cref{fig:loss.curve}). Inputs are the noisy samples (drawn from a standard Gaussian at inference) concatenated channel-wise with the spatial conditions, along with the auxiliary scalar conditions (i.e., noise level and data time deltas). Predicting 69 channels means we have 141 input channels with forcings (\cref{tab:data}). We also find the dynamic time intervals help regularize the model's training dynamics.

Every other layer implements shifting windows in x- and y-directions to increase the effective receptive field. We remove the relative position biases from the scaled cosine attention block and adopt adaptive LayerNorm (adaLN) \citep{perez2018film} and SwiGLU \citep{shazeer2020glu} for the post-normalization layers and fully-connected layers, respectively. Each adaLN normalizes its input tensor modulated by the shared latent embedding $t$, which is the combination of the noise level encoded by a sinusoidal transformation and the embedded predicted time delta, $\delta i$. Two layers with silu nonlinearities create the $t$ embedding. The output decoder layer is a simple linear reshape. 

Specific to this work, the spatial inputs are embedded with a linear projection of $2\times2$ patches. We use a hidden dimension of $1{,}056$, with $12$ attention blocks, each using $12$ attention heads---resulting in a 225M parameter model. Layer biases and modulation weights are zero-initialized; all other weights follow the initialization in \citep{nguyen2024scaling} with a truncated normal distribution (zero mean and standard deviation of $0.02$). Other weight initializations we tested caused our model to diverge during training.

\subsection{Implementation Details} \label{ssec:implementation}

Our models are trained in two phases, pretraining with both the diffusion and consistency model, then finetuning of just our consistency model (\cref{ssec:objectives}). All our models are developed in PyTorch and trained on a cluster of 120$\times$ 64GB Intel Max 1550 GPUs (tiles) for a maximum of 20 million images (15M for pretraining and 5M for finetuning) over 3 days per model; using a local batch size of 1 and distributed data parallelism, this equates to 167K training/update steps. We also use \texttt{bfloat16} in training and \texttt{float32} during inference. 

We had explored many optimizers and learning rate schedules, finding AdamW \citep{loshchilov2017decoupled} to work best with diffusion and finetuning, whereas Muon \citep{jordan2024muon} was the most effective and stable with our consistency model. For AdamW we use a maximum learning rate of $\eta=\text{5e-4}$, $\beta_{1,2} = [0.9, 0.95]$, $\epsilon=1\text{e-}6$, and weight decay $\lambda=1\text{e-}5$ for all parameters except for the positional embedding and modulation layer. With Muon we still use AdamW for 1D parameters and embedding layers with $\eta=3\text{e-}4$, $\beta_{1,2} = [0.9, 0.95]$, $\epsilon=1\text{e-}10$, and $\lambda=0.01$. All 2D transformer weights use $\eta=0.02$, $\lambda=0.01$, and momentum $m=0.95$ with a default of 5 Newton--Schulz iterations. 

Training stability and performance is further improved by a using a cosine learning rate schedule for both optimizers, starting with linear warmup (a minimum value set by scaling $\eta$ by $1\text{e-}4$) for 2M images, cosine annealing until 15M images, and constant $\eta=1\text{e-}5$ during finetuning (\cref{fig:loss.curve}). We also apply an exponential moving average (EMA) with a 500K image halflife and use the EMA parameters during inference. Lastly, while our data (target residuals) is normal in its first two moments, it has variables with heavy tails. We therefore find that a log-uniform noise schedule, $\tau = (1 - u)\,\log(\sigma_{\min}) + u\,\log(\sigma_{\max})$, where $u \sim \mathcal{U}(0,1)$ and noise levels $\sigma_{\min}=0.02$ and $\sigma_{\max}=200$ empirically provide the best results. This only pertains to training as during inference and finetuning, we sample a single step from $t=\tfrac{\pi}{2}$, i.e., the upper bound of $t$ as $\tau \rightarrow +\infty$.

\section{Experimental Results}\label{sec:results}

We first assess global, medium-range forecast skill against baselines (\cref{ssec:results.skill}), then examine long-term stability for subseasonal-to-seasonal forecasting (\cref{ssec:results.stability}), and finally present case studies of extreme events and seasonal trends (\cref{ssec:results.cases}).

\subsection{Global Forecast Skill}\label{ssec:results.skill}

We compare forecast performance with the state-of-the-art data-driven model GenCast \citep{price2023gencast} and a numerical forecasting system (IFS ENS), in addition to our own diffusion model and non-finetuned \name-\textit{B} (base) variant. All medium-range forecasts are made on $\delta i = 6$ intervals with uniformly distributed test samples in 2020. Skill is measured by latitude-weighted ensemble mean root-mean-squared error (RMSE) (\cref{app:fig:skill.rmse}), per-location continuous ranked probability score (CRPS) (\cref{app:fig:skill.crps}), and the model's spread to skill ratio (SSR) (\cref{app:fig:skill.ssr}) relative to target ERA5.

With our diffusion model and limited compute, we generates 12, 15 day forecasts in 7.6 min/forecast with 24 ensembles, using 39 NFEs per autoregressive step with the DPMSolver++ 2S solver \citep{lu2025dpm,karras2022elucidating}, albeit without stochastic churn due to the trigonometric complexity. In contrast, with \name we generate 64 forecasts in 15 sec/forecast with 12 ensembles and only 1 NFE (\textit{30$\times$ empirically faster wall-clock time}). As shown in \cref{fig:skill}, \name remains competitive with the IFS ENS for most variables despite being underdispersive (having an SSR < 1) and only using 12 ensemble members (versus 50 as in GenCast), and finetuning substantially improves stability over \name-\textit{B}. Concretely, \name performs competitively with our diffusion model while running at a fraction of the cost, and it consistently outperforms the pretrained baseline, though it marginally underperforms GenCast. Additional results for other variables are provided in \cref{app:sec:results}. 

We illustrate 7 day forecasts errors in \cref{app:fig:samples.error}, where we see the greatest errors in the ensemble mean nearest the poles across variables. These align with persistent polar forecasting challenges, amplified by higher reanalysis uncertainties in ERA5 (e.g., from observation sparsity) and model uncertainty in our latitude weighted losses. While not qualitatively visible, we find \name's predictions of smoother fields, i.e., geopotential and mean sea level pressure, drift at high zonal wavenumbers to larger scales proportional to lead-time. Other forecast fields maintain accurate spectra and are stable over time (\cref{app:fig:power}). This drift is a known limitation that is addressable with field truncation \cite{lang2024aifs}, and is not inherently detrimental to model performance.

\begin{figure}[!b]
  \begin{subfigure}[t]{0.49\textwidth}
    \centering
    \includegraphics[width=\linewidth]{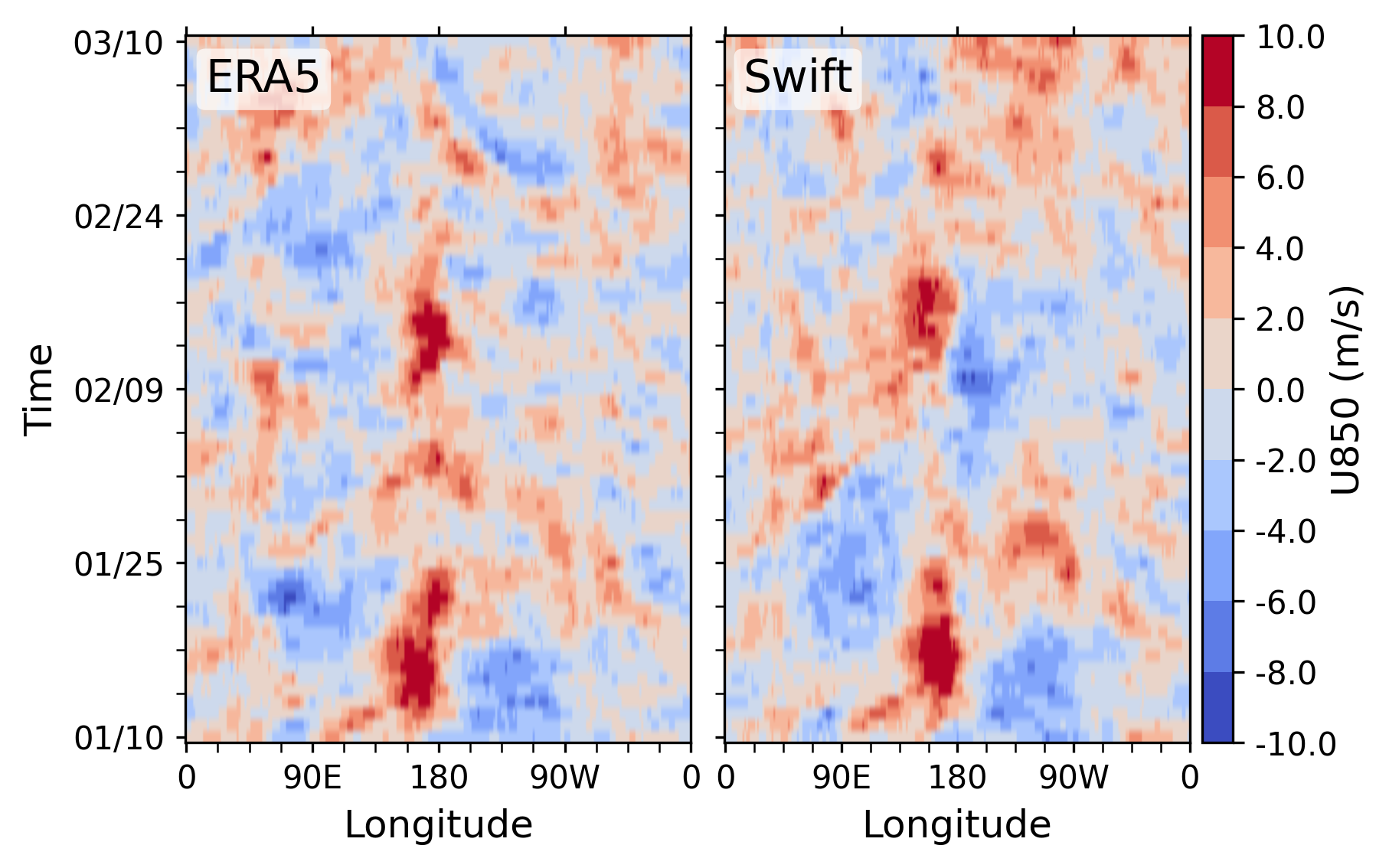}
    \caption{Hovm{\"o}ller diagram}
    \label{fig:s2s:hovmoller}
  \end{subfigure}%
  \hfill
  \begin{subfigure}[t]{0.28\textwidth}
    \centering
    \raisebox{0pt}[0pt][0pt]{%
        \resizebox{\linewidth}{!}{%
        \includegraphics[width=\linewidth]{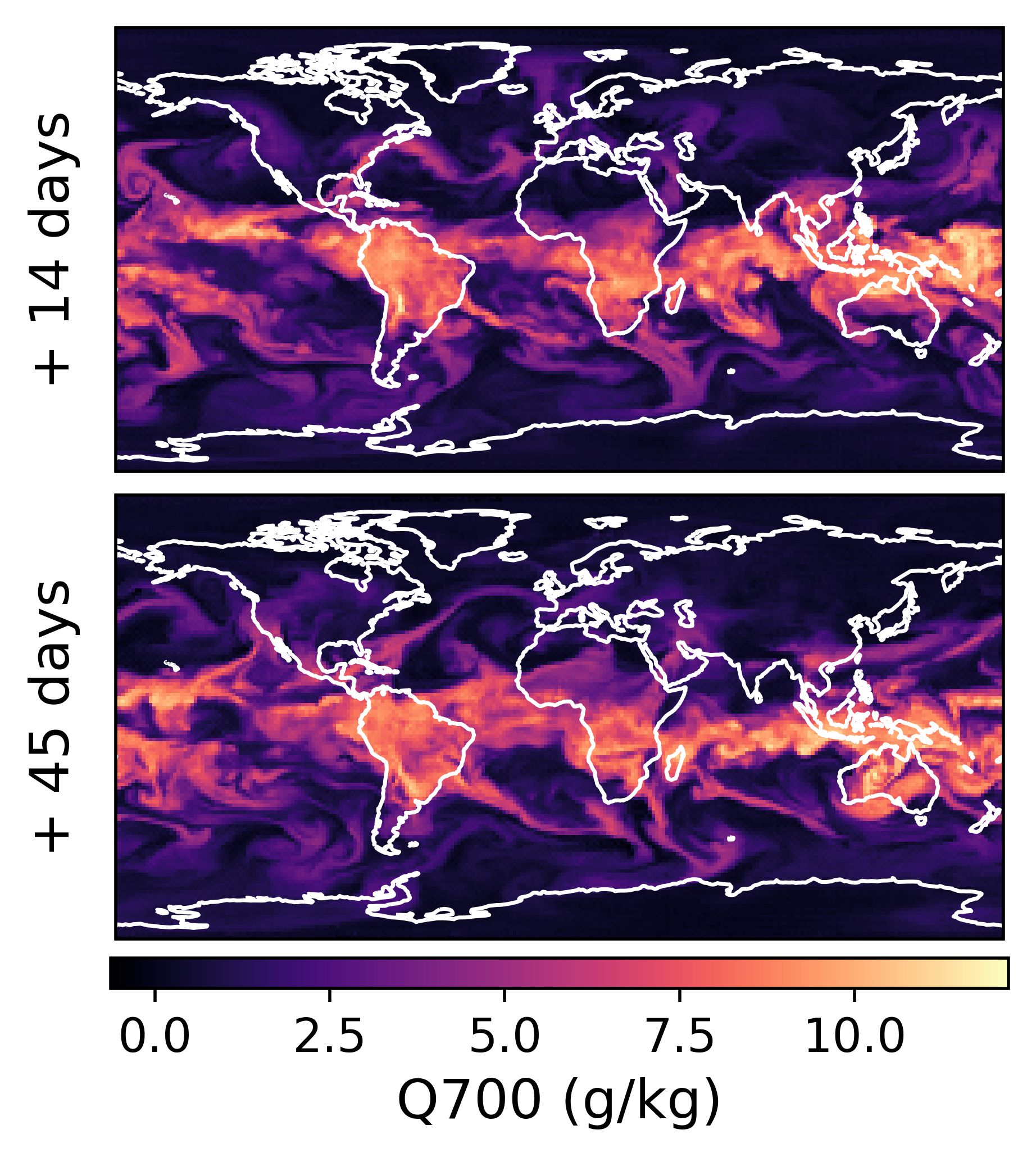}
        }%
    }
    \caption{Global forecast states}
    \label{fig:s2s:forecasts}
  \end{subfigure}%
  \hfill
  \begin{subfigure}[t]{0.224\textwidth}
    \centering
    \includegraphics[width=\linewidth]{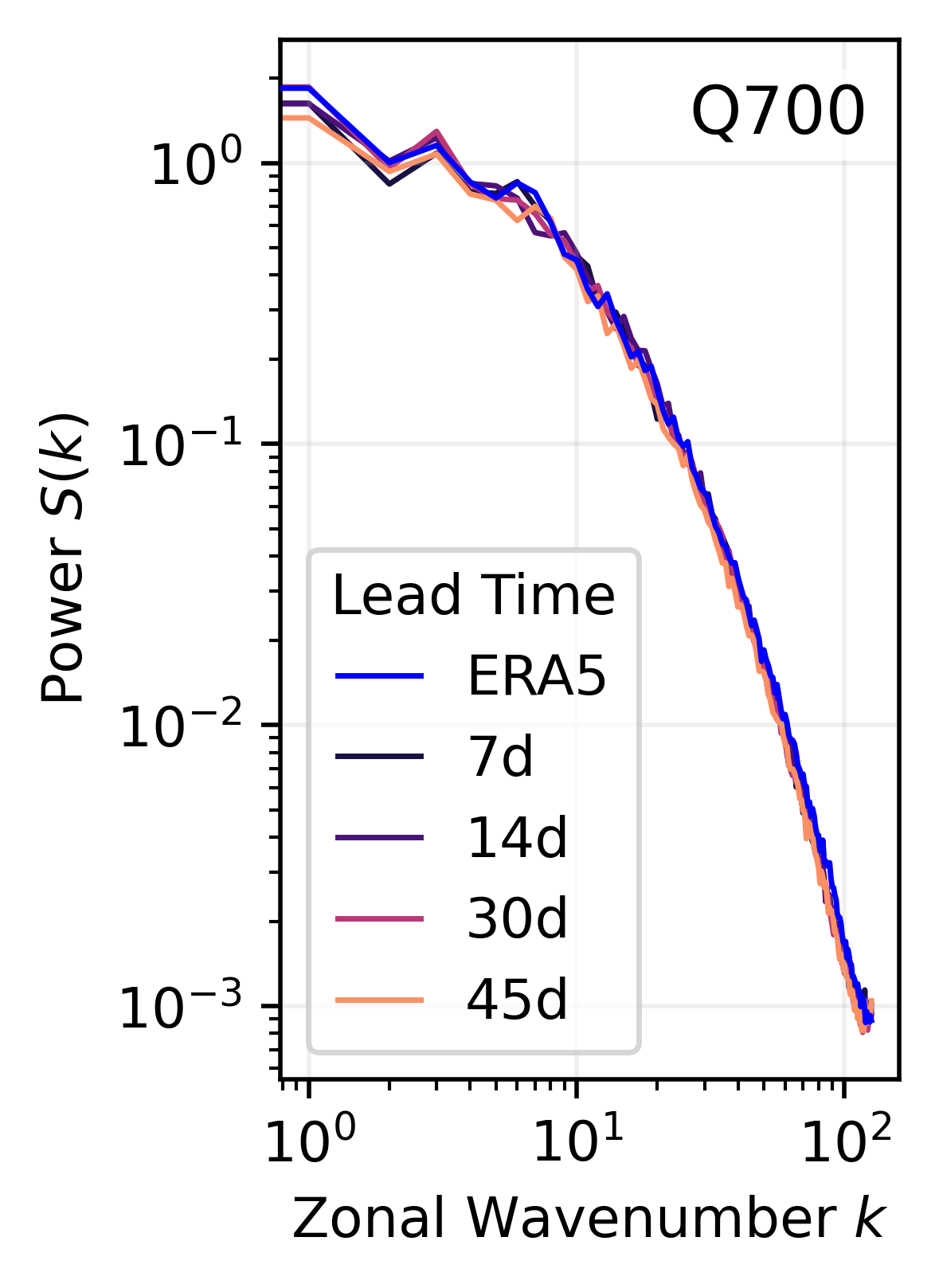}
    \caption{Power spectra}
    \label{fig:s2s:power}
  \end{subfigure}%
  \caption{Long-term stability (with $\delta i = 24$). (a) u850 anomalies (climatology removed) Hovm{\"o}ller \citep{hovmoller1949trough} averaged between 10$^\circ$ N/S; (b) single-member 14- and 45-day predicted fields of q700 initialized on Jan 1, 2020; and (c) power spectra of q700 forecasts averaged over 32 initials and 12 ensembles.}
  \label{fig:s2s}
\end{figure}

\subsection{Long-term Stability}\label{ssec:results.stability}

We evaluate the potential for subseasonal-to-seasonal (S2S) forecasting by assessing long-term stability (qualitatively only to ERA5), which is essential for climate modeling and decision-making under uncertainty. The Hovm{\"o}ller diagram \citep{hovmoller1949trough} in \cref{fig:s2s:hovmoller} shows u850 anomalies from a single-member 60 day forecast, where \name resembles ERA5 by reproducing realistic eastward- and westward-propagating equatorial wave modes, preserving land--ocean boundaries (e.g., around 80$^\circ$ W and 45$^\circ$ E), and correctly disperses high-amplitude equatorial wind events at subseasonal-scales. Additional Hovm{\"o}ller diagrams supporting stability out to 75 days are shown in \cref{app:fig:hovmollers}.

Deterministic data-driven weather models \citep{pathak2022fourcastnet,bi2022pangu,graphcast,nguyen2024scaling}, while computational efficient, tend to blur sharp features at extended lead-times---an artifact of squared-error objectives converging toward the ensemble mean \citep{chattopadhyay2023challenges,ebert2025measuring}, among other influences. In contrast, \name enjoys the efficiency of inference while preserving sharpness, a result of modeling the data distribution as a consistency model during pretraining and calibrating it through finetuning. \cref{fig:s2s:forecasts,fig:s2s:power} illustrates this with global forecasts of q700 and the respective power spectra averaged over 32 initial conditions made on $\delta i=24$ intervals. We find our forecasts retain coherent structures while having realistic power spectra, further supporting stable long-term behavior.

Additional fields on seasonal-scales are visualized in \cref{app:fig:samples}, with a single-member forecast out to 14- and 45-days over all variables. In both cases, the forecasts remain realistically sharp.

\begin{figure}[!ht]
  \begin{subfigure}[t]{0.395\textwidth}
    \centering
    \includegraphics[width=\linewidth]{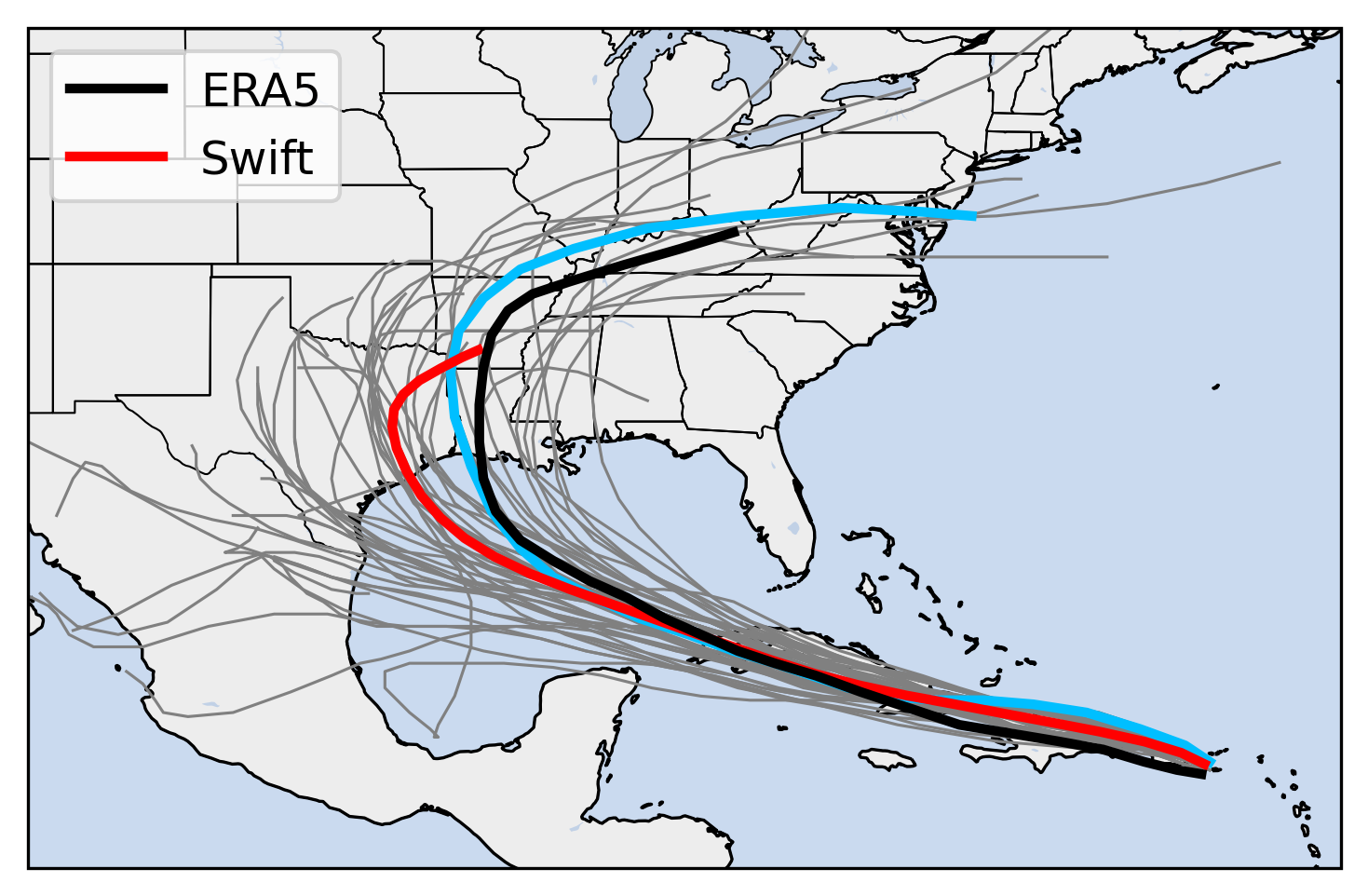}
    \caption{Tropical cyclone tracks}
    \label{fig:ensemble:tc}
  \end{subfigure}%
  \hfill
  \begin{subfigure}[t]{0.6\textwidth}
    \centering
    \includegraphics[width=\linewidth]{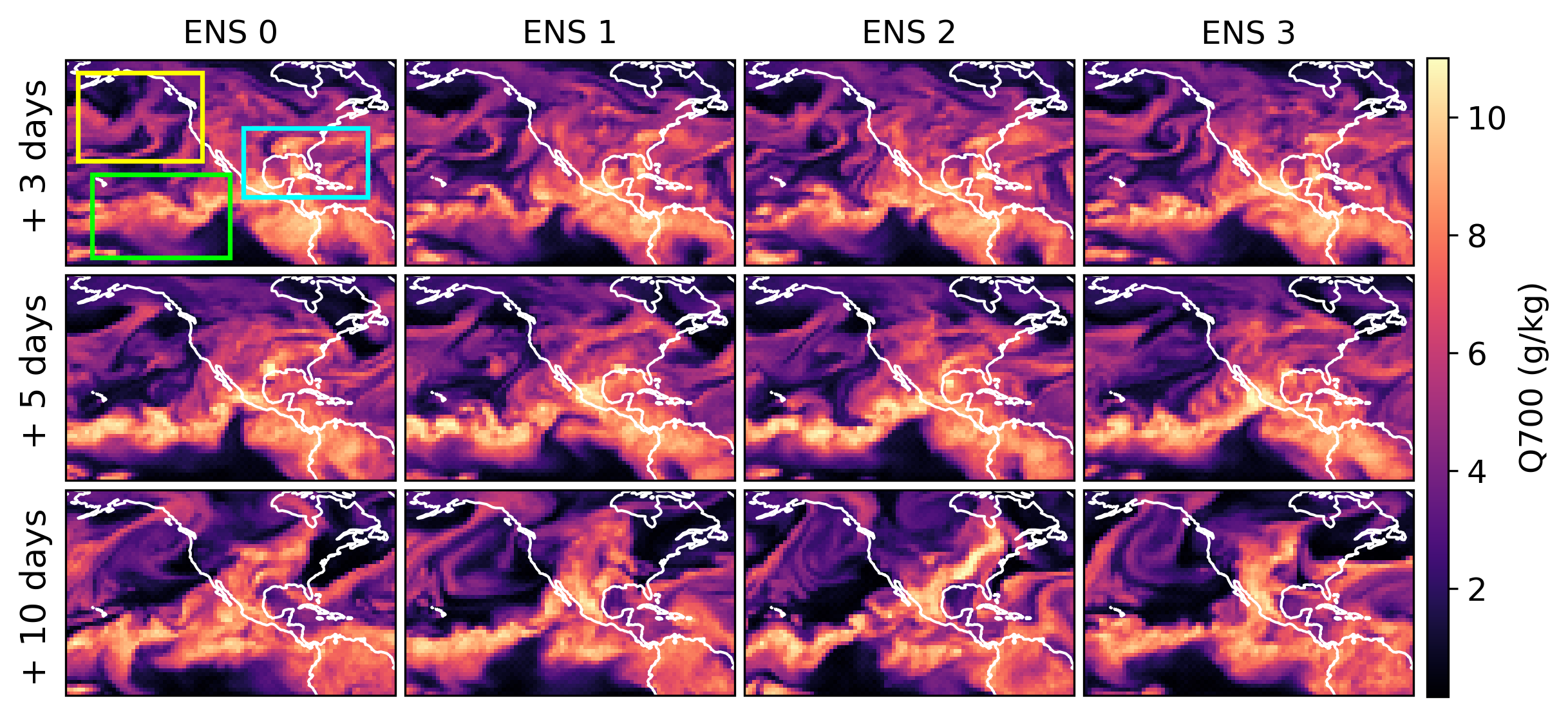}
    \caption{Ensemble forecasts}
    \label{fig:ensemble:forecasts}
  \end{subfigure}%
  \caption{Probabilistic weather forecasting (initialized 2020-08-22T06z; $\delta i=6$). (a) predicted tracks of Hurricane Laura depicted over 48 ensembles; and (b) individual ensembles at increasing lead-time, highlighting areas of a \textcolor{cyan}{tropical cyclone}, an \textcolor{yellow!80!black}{atmospheric river}, and the \textcolor{green!70!black}{intertropical convergence zone}.}
  \label{fig:ensemble}
\end{figure}

\subsection{Case Studies}\label{ssec:results.cases}

While many potential case studies could be examined, we focus on two that highlight \name's ability to: (a) generate diverse and realistic ensembles under extreme weather conditions, and (b) capture seasonal cycles throughout the year from a given initial condition.

\paragraph{Extreme Weather Events.} On August 27, 2020, Hurricane Laura made landfall near Cameron, Louisiana, resulting in \$19 billion in damage and 47 direct deaths \citep{pasch2021hurricane}. Considering its impact, we study this storm and generate 48 forecasts from \name, initialized 5 days before landfall, and superimpose the predicted tracks on the target ERA5 track, computed at 1.4$^\circ$ resolution (\cref{fig:ensemble:tc}). Ensemble uncertainty remains low until the track turns northward, when ensemble spread increases, though several members capture the observed track. Individual realizations over North America are shown in \cref{fig:ensemble:forecasts}, where in q700 we find the tropical cyclone, an atmospheric river in the Gulf of Alaska, and the Intertropical Convergence Zone, consistent with \cite{kochkov2024neural}. These qualitative fields further highlight the realistic atmospheric dynamics and diversity among ensemble members.

\begin{figure}[!t]
    \centering
    \includegraphics[width=\linewidth]{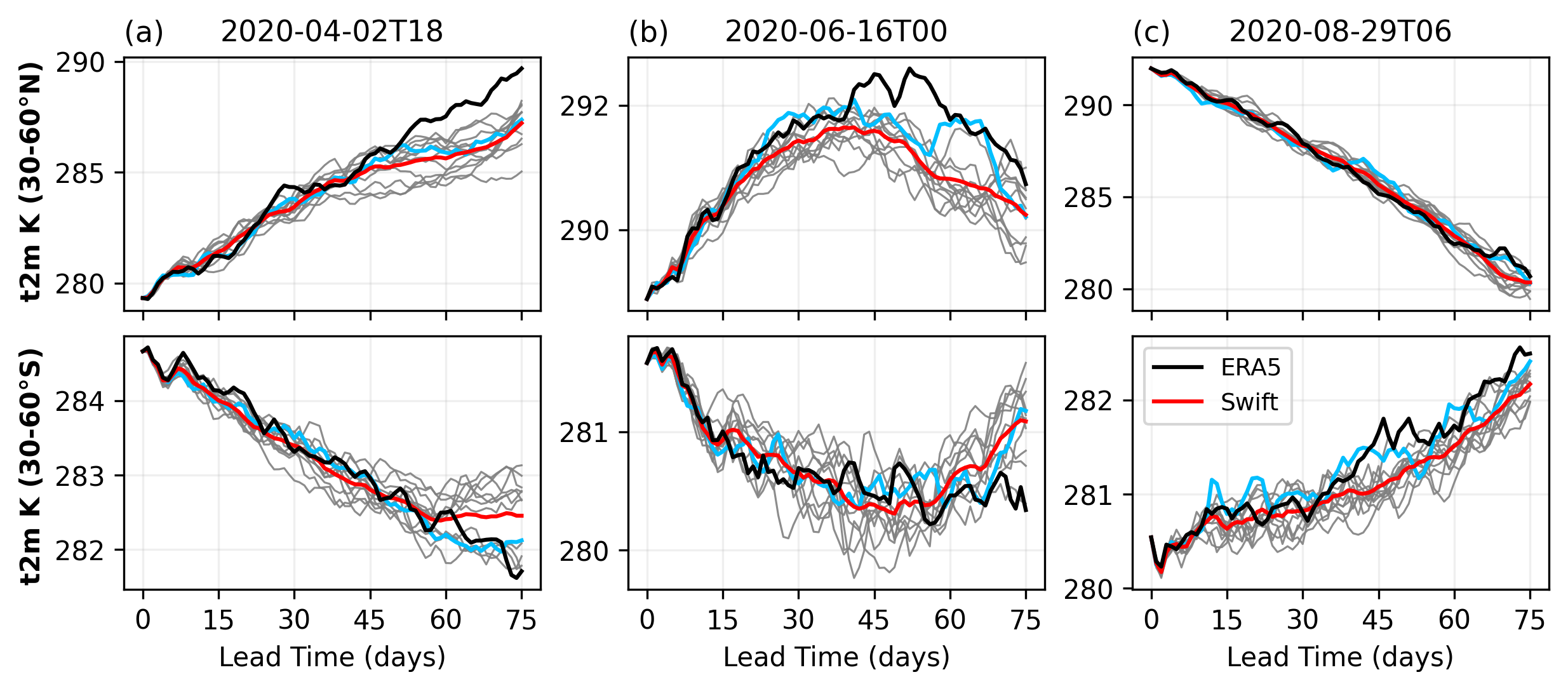}
    \caption{Seasonal cycle forecasts (with $\delta i=24$). Area mean 2 meter temperature for the Northern (top) and Southern (bottom) Hemisphere extratropics for three, seasonally spread initial conditions compared to ERA5 (black) with the \textcolor{cyan}{closest} forecast member.}
    \label{fig:t2m-seasonal}
\end{figure}

\paragraph{Extratropical Seasonal Trends.} Data-driven models often struggle to generate physically consistent behavior, particularly in reproducing slow, forced responses of the atmosphere rather than merely propagating variability. To assess this, we initialize \name from several distinct initial conditions over 2020 and examine the evolution of extratropical 2 meter temperature over 75 day periods (\cref{fig:t2m-seasonal}). Both individual ensemble members and the mean follow the expected response to the changing in the forcing conditions on seasonal time-scales. That is, the Northern Hemisphere warms through boreal spring and cools into autumn, while the Southern Hemisphere exhibits the opposite progression. The seasonal transition initialized in June is also captured in the rollout. Across initials, \name remains stable and phase-aligned with ERA5, though with a tendency to slightly understate the amplitude. 

\section{Conclusion}

Our analyses demonstrate that our generative, autoregressively finetuned consistency model, \name, achieves skillful 6-hourly forecasts that remain stable for up to 75 days while being significantly more computationally efficient than diffusion baselines (up to 39$\times$ faster). Although more exploration is needed, preliminary results suggest that \name has the potential to improve subseasonal-to-seasonal (S2S) forecasting. Case studies further demonstrate the ability to capture diverse, realistic ensembles during extreme events and model seasonal cycles. Taken together, these results mark a significant step toward skillful machine learning-based ensemble weather forecasting that spans medium-range to seasonal-scales, without requiring substantially more compute compared to deterministic models.

Despite these achievements, several limitations remain. Comparisons used a reduced number of initial conditions and ensemble members due to computational and storage constraints; we expect improved performance by scaling both. Additionally, forecast results are underdispersive with finetuning leading to better long term spread at the expense of short term SSR (days 0-4), which could be addressed with a better learning rate schedule during finetuning. Lastly, our medium-range and long-term forecasts used $\delta i=6$- and $\delta i=24$-hourly intervals, respectively, leaving the potential to improve stability at a higher temporal resolution with a Pangu-Weather style \citep{bi2022pangu} greedy schedule. 

Looking ahead, we aim to explore consistency distillation with a larger, high-resolution model (e.g., \cite{hatanpaa2025aeris}) to make operational deployment more accessible. Moreover, we can leverage methods such as classifier-free guidance to study improvements to forecast performance \citep{ho2022classifier}. Finally, considering our method is agnostic to architecture, it would be insightful to benchmark alternative backbones.

\section*{Acknowledgments}
This work has been supported by the U.S Department of Energy (DOE) Office of Cybersecurity, Energy Security, and Emergency Response (CESER) and also by the Laboratory Directed Research and Development (LDRD) Program at Argonne National Laboratory through the U.S. Department of Energy (DOE) contract DE-AC02-06CH11357. Computing resources come from the Argonne Leadership Computing Facility, a U.S. Department of Energy (DOE) Office of Science user facility at Argonne National Laboratory.

\bibliographystyle{unsrt} 
\bibliography{main}

\newpage
\appendix
\section*{Appendix}

\section{Full Field Visualizations}

\begin{figure}[!ht]
  \centering
  \includegraphics[width=\linewidth]{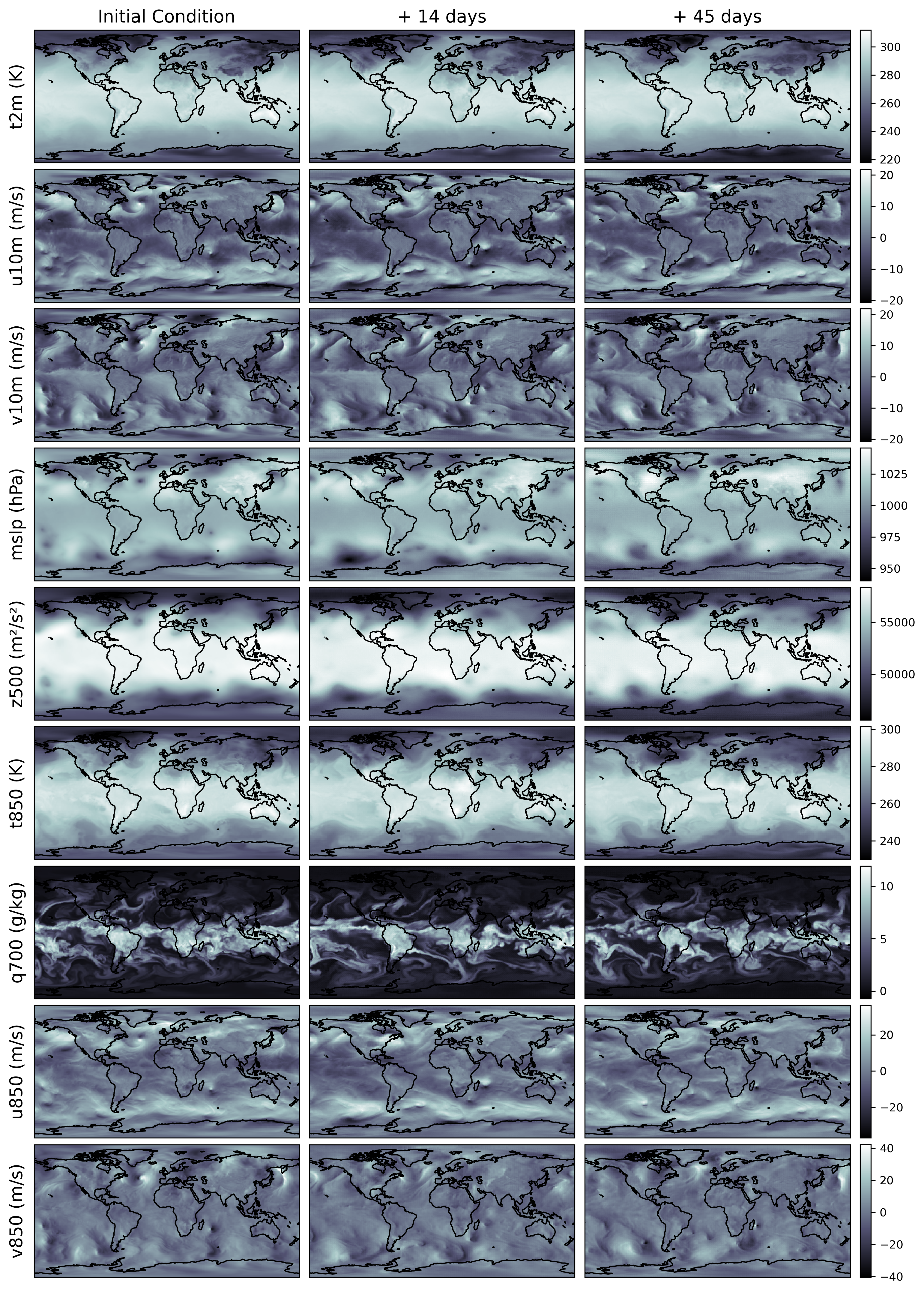}
  \caption{Additional samples from a single ensemble member (initialized on 01-01-2020T0z).}
  \label{app:fig:samples}
\end{figure}

\begin{figure}[!ht]
  \centering
  \includegraphics[width=\linewidth]{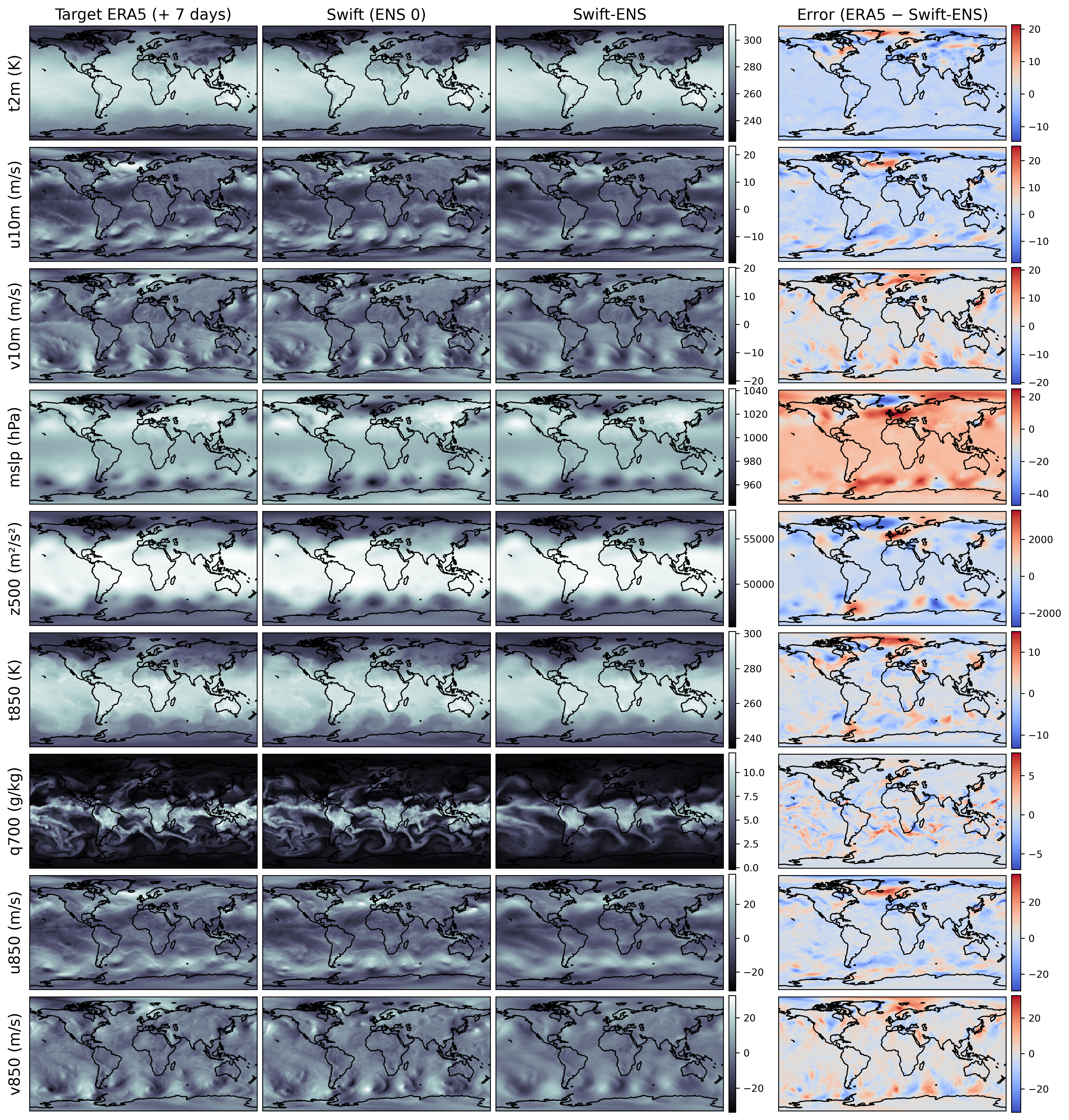}
  \caption{Forecast error at 7-day lead-time (initialized on 01-01-2020T0z).}
  \label{app:fig:samples.error}
\end{figure}

\clearpage
\section{Additional Medium-Range Results}\label{app:sec:results}

\begin{figure}[!ht]
    \centering
    \includegraphics[width=\linewidth]{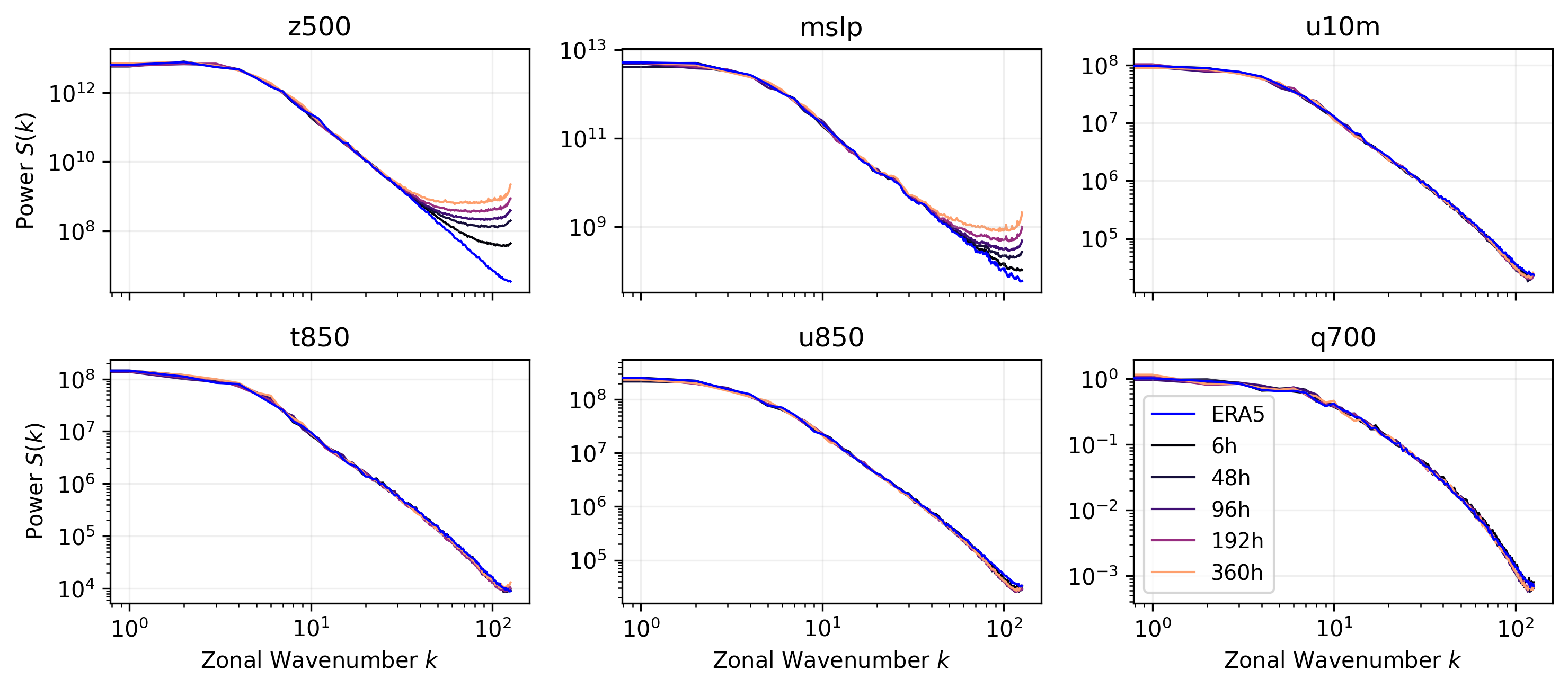}
    \caption{Power spectra compared to target ERA5 from generated ($\delta i=6$) 15 day forecasts, averaged over 32 initial conditions and 12 ensembles shown for each variable between 6--360h.}
    \label{app:fig:power}
\end{figure}

\begin{figure}[!ht]
  \centering
  \includegraphics[width=\linewidth]{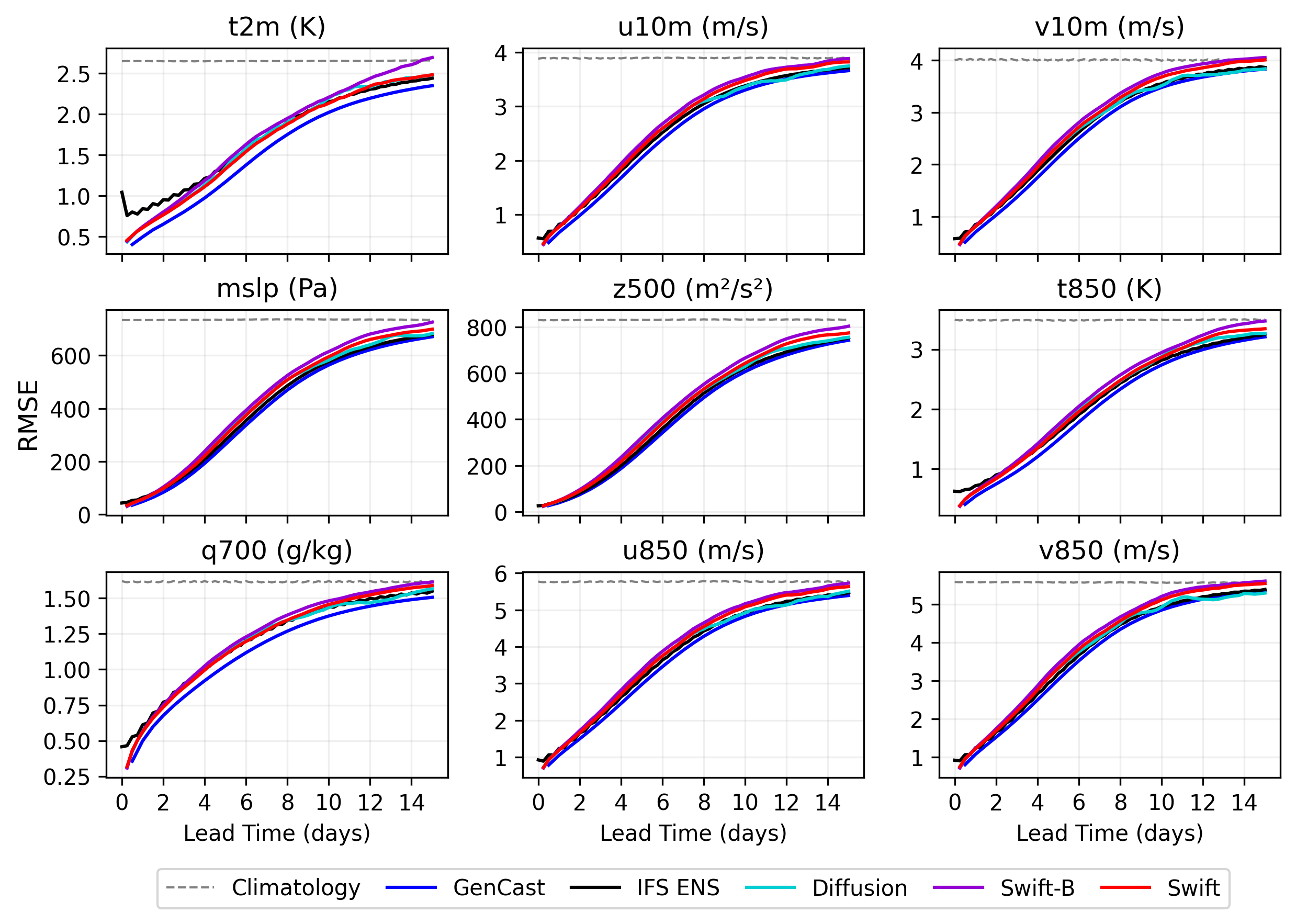}
  \caption{Latitude-weighted ensemble mean root-mean-squared error (RMSE).}
  \label{app:fig:skill.rmse}
\end{figure}

\begin{figure}[!ht]
  \centering
  \includegraphics[width=\linewidth]{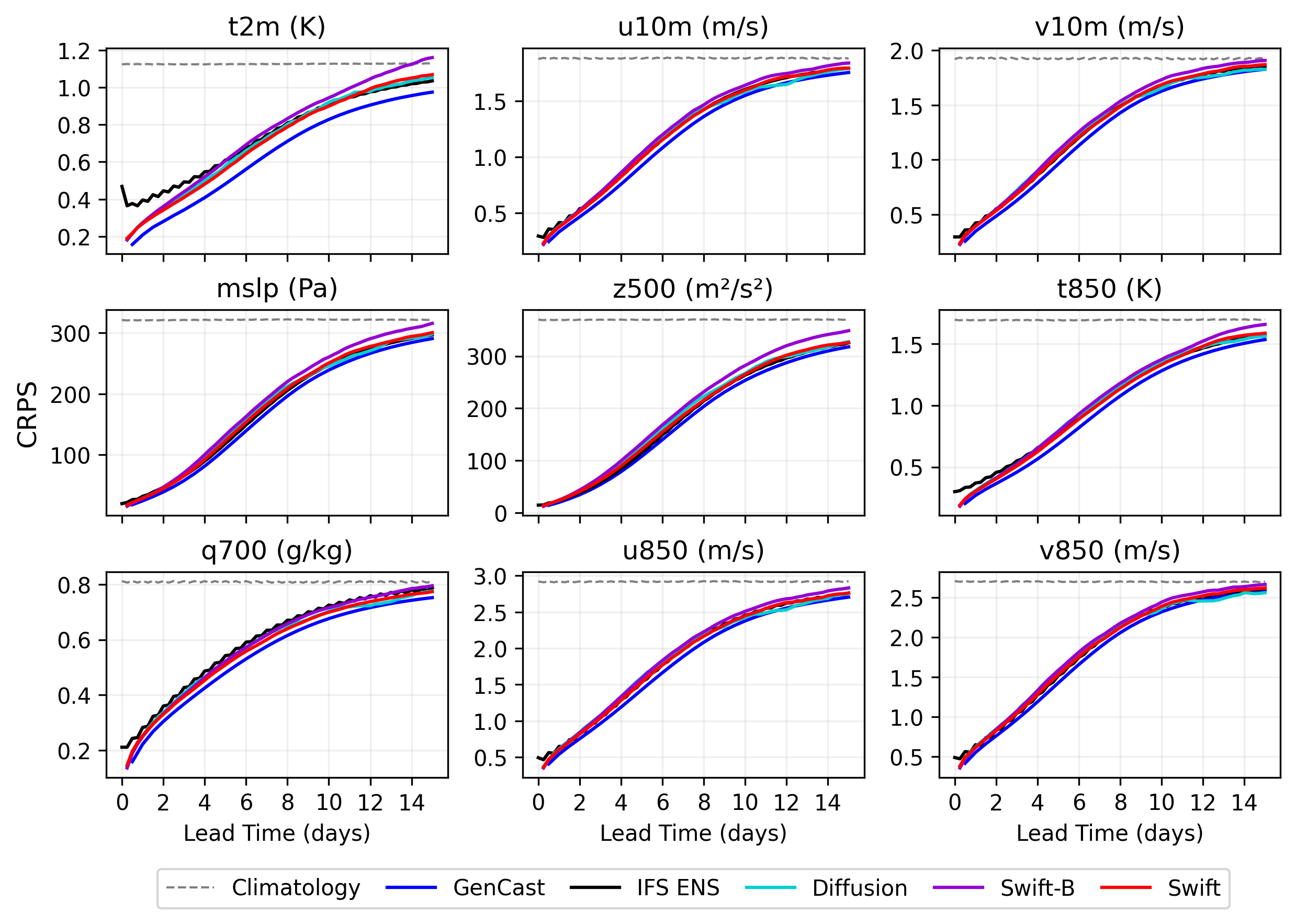}
  \caption{Latitude-weighted continuous ranked probability score (CRPS).}
  \label{app:fig:skill.crps}
\end{figure}

\begin{figure}[!ht]
  \centering
  \includegraphics[width=\linewidth]{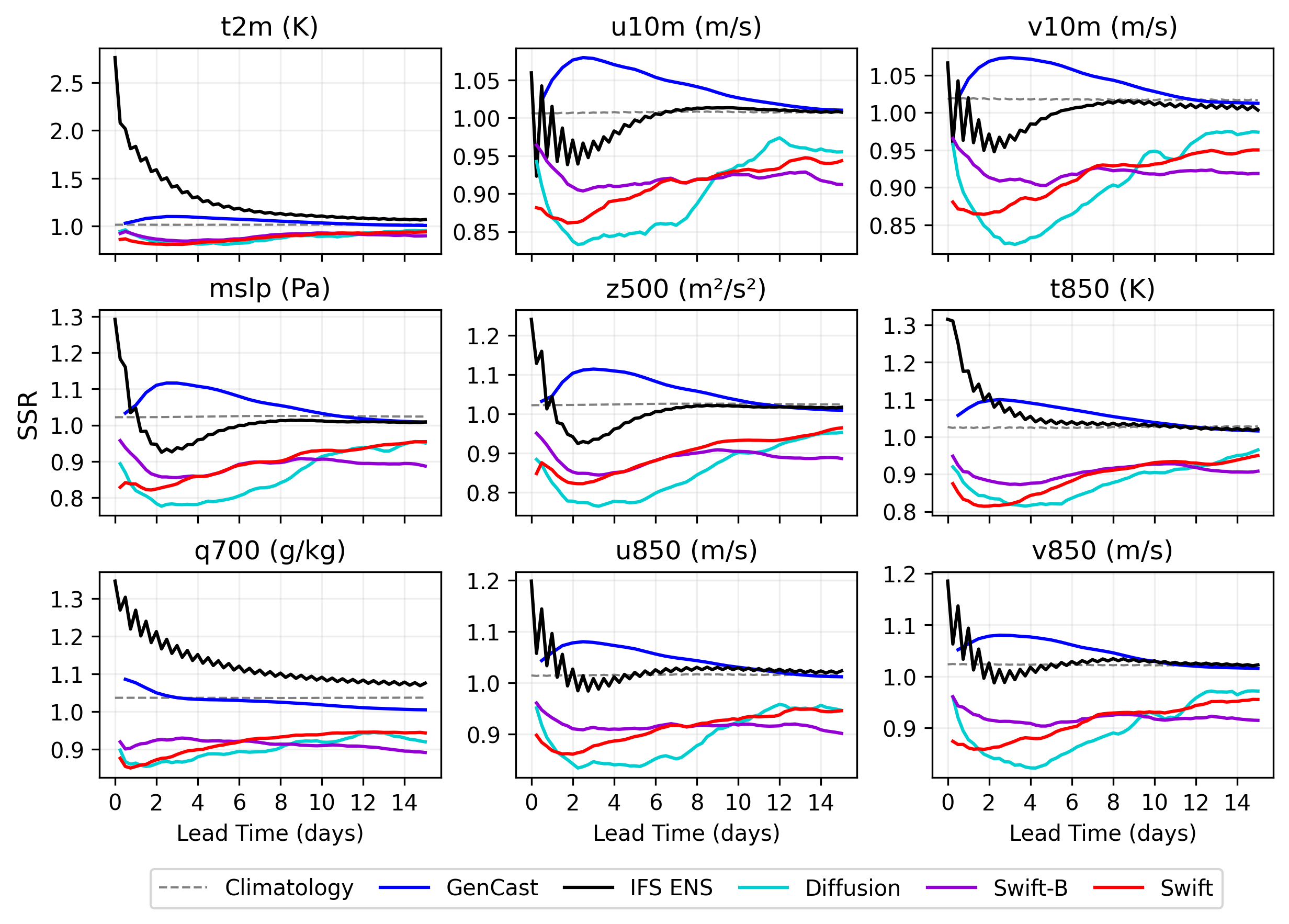}
  \caption{Latitude-weighted spread to skill ratio (SSR).}
  \label{app:fig:skill.ssr}
\end{figure}

\clearpage
\section{Additional Hovm{\"o}ller Diagrams}

\begin{figure}[htbp]
    \centering
    \begin{subfigure}{0.49\textwidth}
        \centering
        \includegraphics[width=\linewidth]{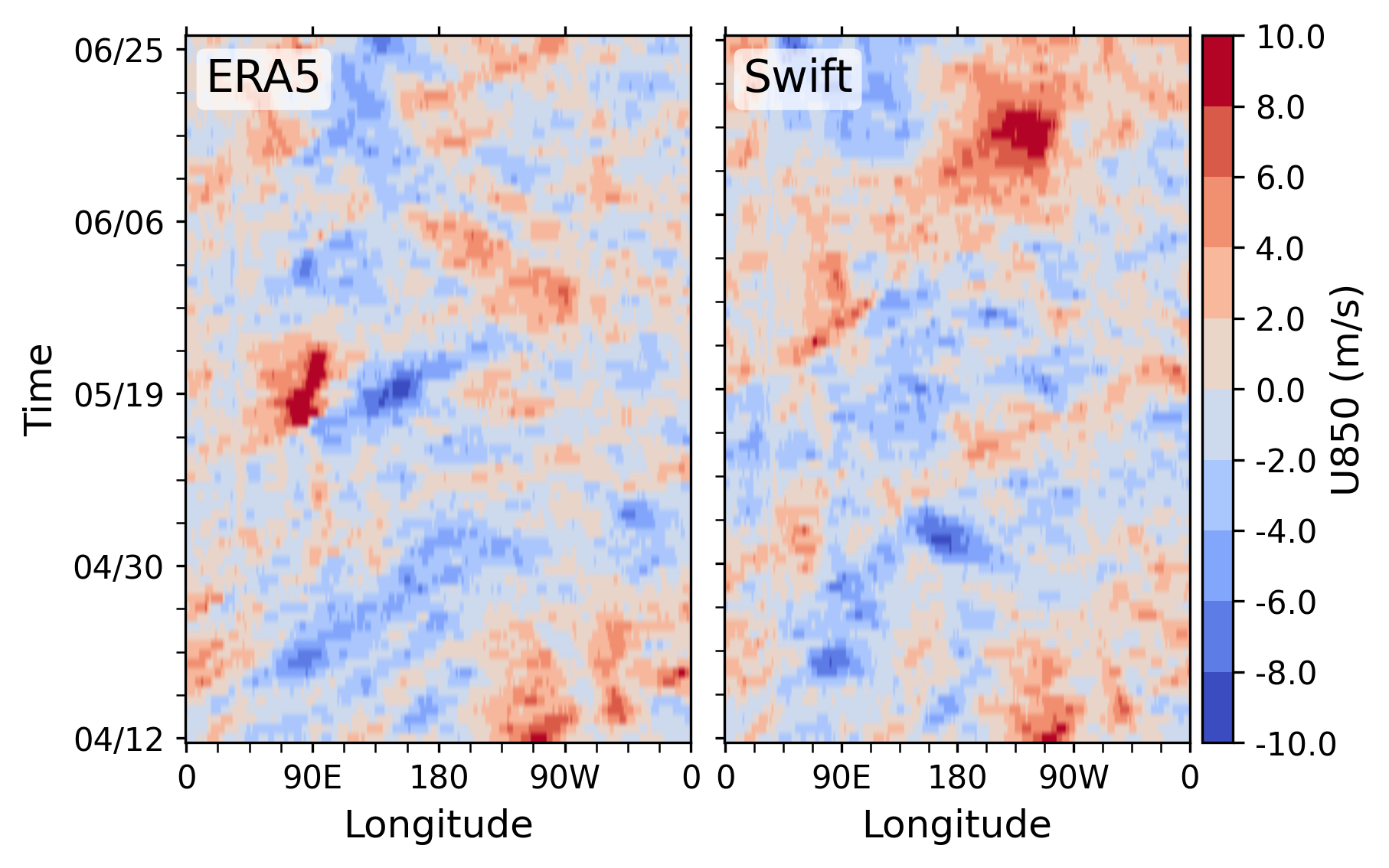}
        \caption{2020-04-12T0z}
        \label{fig:a}
    \end{subfigure}
    \hfill
    \begin{subfigure}{0.49\textwidth}
        \centering
        \includegraphics[width=\linewidth]{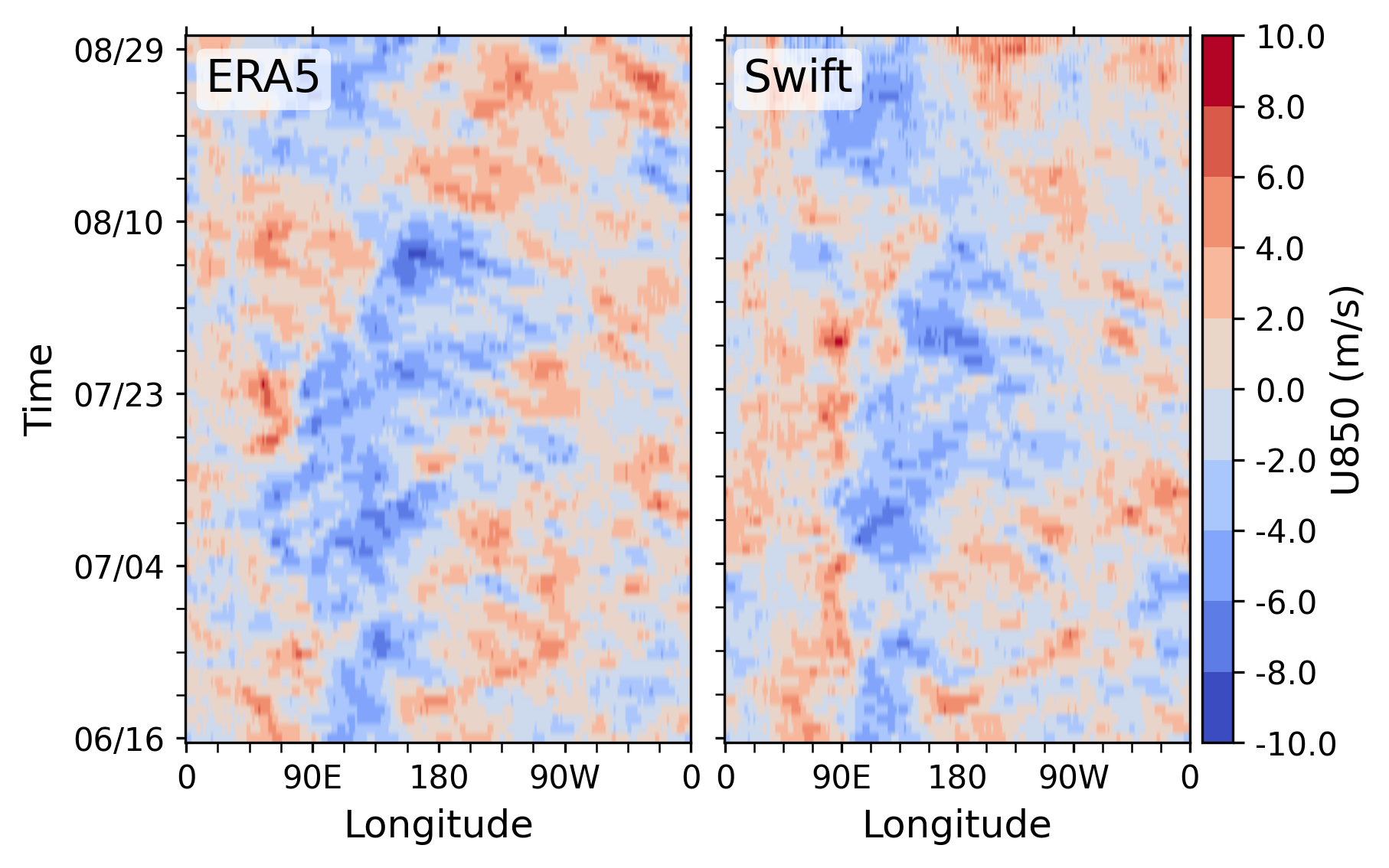}
        \caption{2020-06-16Tz}
        \label{fig:b}
    \end{subfigure}

    \vskip\baselineskip

    \begin{subfigure}{0.49\textwidth}
        \centering
        \includegraphics[width=\linewidth]{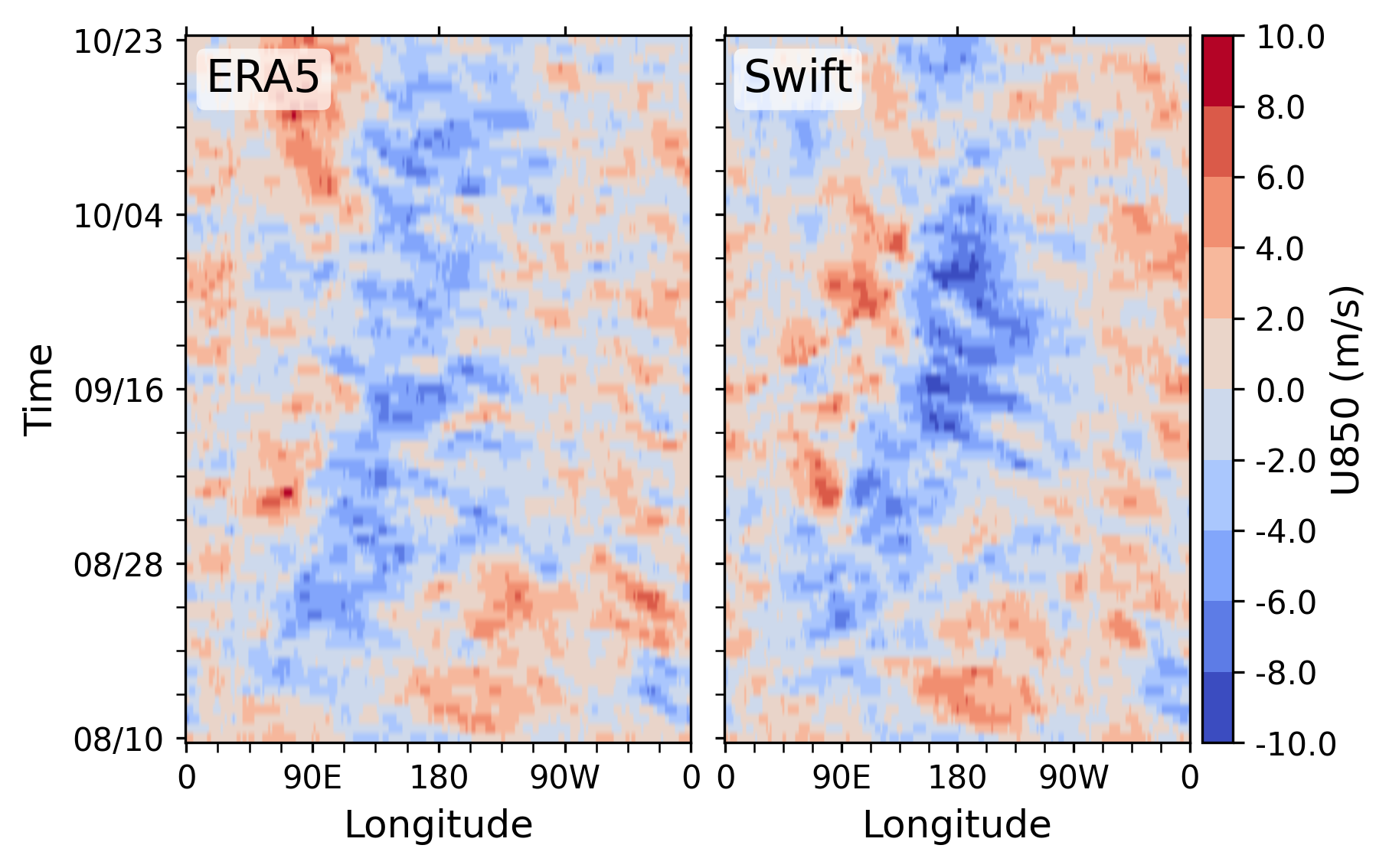}
        \caption{2020-08-10T18z}
        \label{fig:c}
    \end{subfigure}
    \hfill
    \begin{subfigure}{0.49\textwidth}
        \centering
        \includegraphics[width=\linewidth]{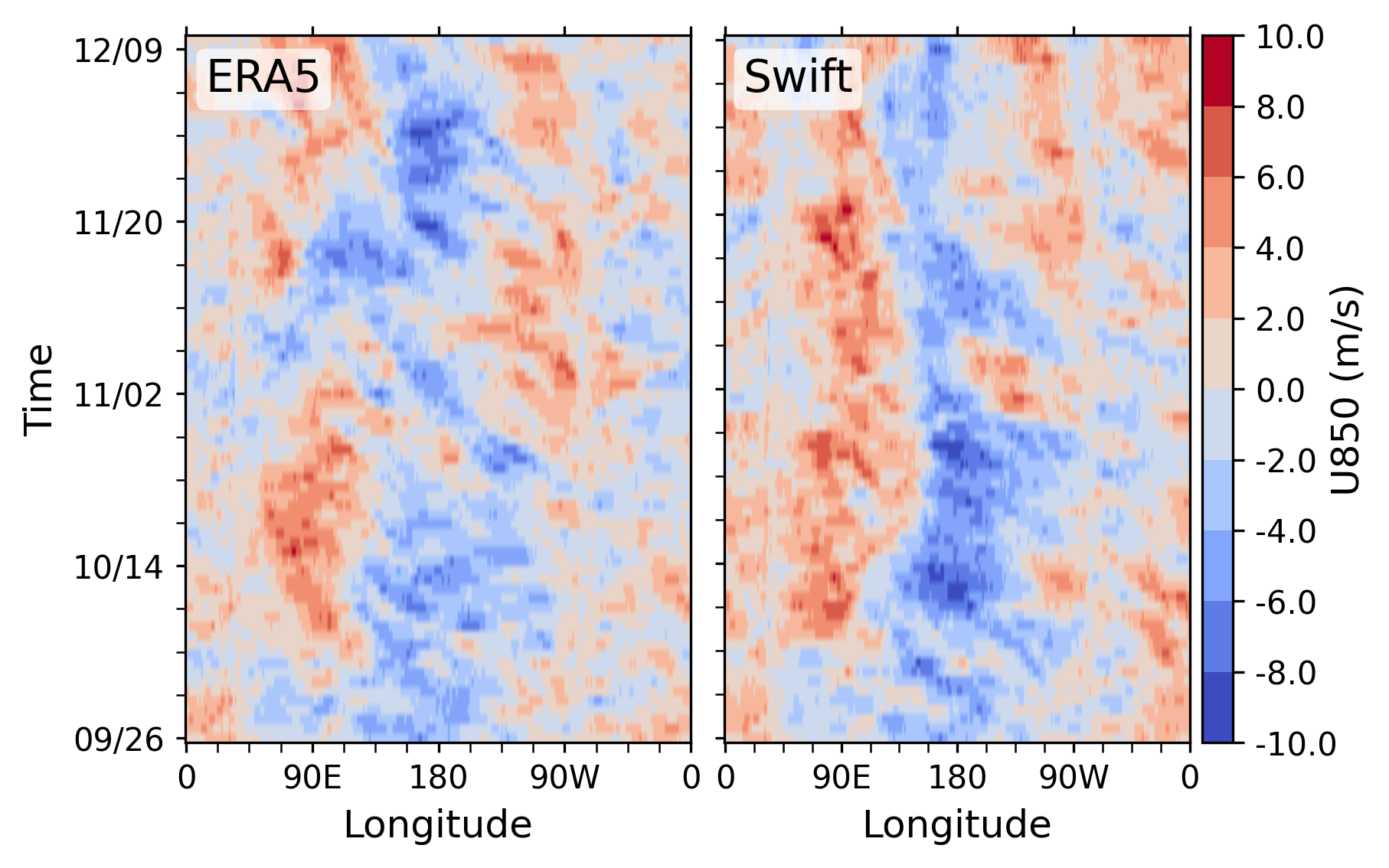}
        \caption{2020-09-26Tz}
        \label{fig:d}
    \end{subfigure}

    \caption{Additional u850 anomaly (climatology removed) Hovm{\"o}ller diagrams highlighting the tropics with averages between 10$^\circ$ N/S over 75 days with initials over the year in (a)--(d).}
    \label{app:fig:hovmollers}
\end{figure}

\end{document}